\documentclass{article}

\makeatletter
\def\input@path{{media/}}
\makeatother
\usepackage{PRIMEarxiv}

\usepackage[utf8]{inputenc}
\usepackage[T1]{fontenc}
\usepackage{hyperref}
\usepackage{url}
\usepackage{booktabs}
\usepackage{amsmath}
\usepackage{amssymb}
\usepackage{amsfonts}
\usepackage{microtype}
\usepackage{graphicx}
\usepackage{array}
\usepackage{tabularx}
\usepackage[numbers,sort&compress]{natbib}
\usepackage[labelfont=bf]{caption} 

\newcommand{\pkg}[1]{\texttt{#1}}

\title{Cross-domain benchmarks reveal when coordinated AI agents improve scientific inference from partial evidence}

\author{%
\paperauthors{Fiona Y. Wang$^{1,2}$ \quad Markus J. Buehler$^{1,3,4,5,*}$}\\[0.25em]
\paperaffiliations{%
$^{1}$Laboratory for Atomistic and Molecular Mechanics (LAMM)\\
$^{2}$Department of Biological Engineering\\
$^{3}$Department of Mechanical Engineering\\
$^{4}$Department of Civil and Environmental Engineering\\
$^{5}$Center for Computational Science and Engineering, Schwarzman College of Computing\\
Massachusetts Institute of Technology, 77 Massachusetts Avenue, Cambridge, MA 02139, USA\\
$^{*}$Corresponding author: mbuehler@mit.edu%
}%
}
\date{}

\begin{document}
\maketitle

\begin{abstract}
Scientific evidence often spans instruments, databases, and disciplines, so no single source records the full phenomenon. This makes it difficult to determine when coordinated AI agents add value over simpler scientific workflows. We evaluate this question with a cross-domain benchmark spanning four scientific tasks: mapping molecular structure into musical representations, detecting historical paradigm shifts in science, identifying vector-borne disease emergence, and vetting transiting-exoplanet candidates. Each case uses a frozen evaluation panel, predefined scoring protocols, explicit baselines, ablations or null controls, and stated limitations. The results define three operating regimes. When different disciplines each capture only part of the phenomenon, cross-channel composites improve over single-channel baselines: climate-vector emergence reaches AUROC 0.944 and exoplanet vetting reaches AUROC 0.955. However, the exoplanet workflow is effectively tied with a strong combined-summary baseline, showing that decomposition does not always improve top-line performance. When one signal dominates, as in paradigm-shift detection, coordination mainly improves interpretation and traceability. For molecular sonification, the gain is representational rather than predictive. ScienceClaw $\times$ Infinite provides the auditable artifact and provenance layer for this evaluation. The benchmark therefore assigns value to coordination only when the corresponding performance, provenance, or representation claim is supported by explicit comparators.
\end{abstract}


\section{Introduction}

Autonomous AI systems for scientific discovery are proliferating~\cite{lu2024ai,ghafarollahi2025sciagents,gottweis2025towards,aygun2026software,ghareeb2025robin,buehler2024graph_reasoning,stewart2025molecular, wang2025swarmslargelanguagemodel, ghafarollahi2025sparksmultiagentartificialintelligence, ghafarollahi2024protagents, stewart2026graphagentsknowledgegraphguidedagentic}, yet evaluations often remain demonstrations or narrow task benchmarks~\cite{wang2023scientific,berens2023ai}. For cross-domain science, the sharper question is when artifact-mediated, coordinated scientific agents actually change what can be concluded. That change may appear as better discrimination on a fixed panel, earlier detection from lead-time evidence, recovery of structure across domains, or a clearer account of where the evidence stops. It may also be absent, in which case coordination mainly improves interpretation and provenance, or a simpler baseline is sufficient. The primary need is therefore a benchmark framework that can compare coordinated workflows with explicit alternatives across domains.

We distinguish three claims that are often conflated in agentic-science evaluations: performance improvement, interpretability and provenance improvement, and representational transformation. The benchmark credits coordination only when the corresponding claim is supported relative to explicit comparators.

Many cross-domain settings are distributed across incomplete observational systems. Here, distributed evidence means that different disciplines, instruments, databases, or communities each observe only part of the phenomenon, so no single channel is sufficient to support the full inference.
Climate data can indicate vector suitability without proving establishment or transmission~\cite{lafferty2009ecology,caminade2014impact,medlock2015vectors,semenza2022climate,kraemer2019spread}. Ecological records can show vector presence without human disease. Epidemiological reports often arrive after the biological risk has formed. Similar fragmentation appears in exoplanet vetting, where transit shape, stellar context, archival cross-checks, and follow-up confirmation are produced by different instruments and communities~\cite{thompson2018kepler_dr25,coughlin2016robovetter,morton2016falsepositive}. It also appears in molecular sonification and meta-science, where cross-domain representations or bibliometric signals must be interpreted across incompatible feature spaces~\cite{buehler2022attention,rdkit,music21,k1962}. In these settings, coordination is a testable hypothesis: preserving intermediate evidence and moving it across disciplinary tools may change the conclusion, or it may only change how the conclusion is documented.

We therefore evaluate coordinated scientific agents as benchmarkable scientific workflows rather than as demonstrations of agent behavior. The primary contribution is a cross-domain benchmark framework: each application uses a frozen panel, predefined scoring protocols, explicit baselines, a single-agent or summary comparator where appropriate, ablations or null controls, and a limitation statement. We instantiate the framework in four domain-spanning applications: molecular sonification, retrospective detection of scientific paradigm shifts, vector-borne disease emergence, and transiting-exoplanet vetting~\cite{buehler2022attention,k1962,nash2001outbreak,ryan2019global,quintana2014kepler186,thompson2018kepler_dr25}. The cross-application comparison yields a regime map identifying where coordination matters: distributed-evidence settings, where coordination improves discrimination over single-channel baselines; dominant-channel settings, where coordination mainly improves interpretation, provenance, and auditability; and representational-mapping settings, where coordination changes the object of inference by recovering cross-domain structure rather than improving prediction.

ScienceClaw~$\times$~Infinite (\url{https://lamm.mit.edu/infinite/}) provides the supporting infrastructure for this evaluation. The coordination layer connects domain tools, specialist agents, and content-addressed artifacts; the public-record layer links artifacts to narrative synthesis and makes each run traceable beyond a local execution log. This substrate makes the benchmark auditable by preserving intermediate artifacts, provenance, reuse, and public investigation records. The work builds on prior efforts connecting materials representations, domain-adapted language models, graph reasoning, literature mining, and multi-agent scientific design~\cite{buehler2022attention,luu2024bioinspired,buehler2024graph_reasoning,ghafarollahi2025sciagents,lu2025finetuning,buehler2025preflexor,yang2025peptide,stewart2025molecular,ghafarollahi2024atomagents}. The companion systems paper describes the coordination mechanism, artifact model, and runtime architecture~\cite{wang2026scienceclaw_infinite}.

\section{Results}

\subsection{A benchmark framework for coordinated scientific agents}

We introduce a cross-domain benchmark framework for evaluating coordinated scientific agents. The framework is designed to answer a specific question: when does artifact-mediated coordination change the supported scientific claim relative to simpler alternatives? Each application therefore pairs a coordinated-agent workflow with a frozen evaluation panel, predefined scoring protocols, scripted baselines, a single-agent or summary comparator where appropriate, ablations or null controls, and an explicit limitation statement. Coordination is credited only when it changes the supported performance, provenance, representation, or inference claim relative to these comparators.

Operationally, we use the term coordinated scientific agent workflow to denote a workflow in which domain-specialist agents, models, or tools produce typed, content-addressed intermediate artifacts that are consumed by downstream scoring or synthesis steps. A single-agent summary baseline may access the same extracted channel flags or summary features, but it does not preserve channel-specific artifact exchange, provenance, or reuse as part of the inference process.

The framework is deliberately portable across scientific tasks that do not share data types or endpoints. In Sound of Molecules, the frozen panel is a 16-compound manifest scored for retrieval, same-class nearest-neighbor coherence, and robustness against chemical baselines and shuffled-label controls. In Computational Kuhn, it is a 16-shift versus 16-control retrospective panel with predeclared recognition dates, citation/semantic/funding ablations, and simpler bibliometric comparators. In Climate-Vector Emergence, it is a 12-event versus 12-control matched panel with literature-anchored climate, ecological, and epidemiological first-signal years. In Cosmic Filter, it is a 12-confirmed versus 12-false-positive exoplanet panel scored from transit-shape, stellar-context, archival, and follow-up evidence.

Across these four applications, the benchmark instantiates 19 comparator or ablation arms, 11 controls or null tests, and 28 reported metrics (Table~\ref{tab:case_summary}). These include retrieval@3 and nearest-neighbor coherence for structure recovery, leave-one-pair-out AUROC and matched-pair accuracy for discrimination, lead time for early warning, permutation tests for null behavior, and calibration or robustness checks where appropriate. The point of the framework is not to impose one universal metric. It is to make coordinated-agent claims auditable across domains by requiring the same evidential components: frozen panels, explicit baselines, ablations, nulls, and stated limits.

The comparator design is central to the benchmark. Sound of Molecules is tested against Morgan-fingerprint, physicochemical, no-3D, shuffled-label, and random-mapping alternatives, which separates structure recovery from aesthetic analogy. Computational Kuhn includes an equal-weight summary baseline, a scripted bibliometrics baseline, and citation-only, semantic-only, and funding-only ablations, which exposes the dominant citation channel. Climate-Vector Emergence compares the coordinated composite with key alternatives including climate-only, ecology-only, epidemiology-only, and combined-fraction scoring, which tests whether lead-time-weighted cross-domain evidence adds signal. Cosmic Filter compares the four-channel composite with transit-only, shape-plus-stellar, no-archival, no-follow-up, and single-agent combined-fraction alternatives, which distinguishes single-signal vetting gains from gains due to explicit evidence decomposition. Table~\ref{tab:case_summary} therefore serves as the framework table, showing how a common benchmark structure applies across scientific domains with different data types, endpoints, and failure modes.

\begin{table*}[hbtp]
\centering
\small
\caption{{\bf Cross-domain benchmark framework for evaluating coordinated scientific agents.} Each application is specified by a scientific question, frozen panel, scoring rule, comparator set, primary result, and key limitation. This shared structure makes coordinated-agent claims comparable across domains.}
\label{tab:case_summary}
\setlength{\tabcolsep}{4pt}
\begin{tabularx}{\textwidth}{@{}>{\raggedright\arraybackslash}p{0.13\textwidth} >{\raggedright\arraybackslash}p{0.16\textwidth} >{\raggedright\arraybackslash}p{0.18\textwidth} >{\raggedright\arraybackslash}X >{\raggedright\arraybackslash}X >{\raggedright\arraybackslash}p{0.16\textwidth}@{}}
\toprule
Application & Cross-domain question & Panel and scoring & Comparators & Primary result & Key limitation \\
\midrule
Sound of Molecules & Does a descriptor-to-harmony mapping recover chemical structure? & Fixed 16-compound panel; retrieval@3, same-class nearest neighbors, robustness & Morgan fingerprints, physicochemical cosine, shuffled labels, random mappings, no-3D ablation & Retrieval@3: 0.2708; same-class NN: 0.6875; robustness: 0.8021 & Post-hoc aesthetic interpretation risk; small panel \\
Computational Kuhn & Can summary signals discriminate historical paradigm shifts? & 16 shifts and 16 matched controls; leave-one-pair-out AUROC and lead time & Scripted bibliometrics, equal-weight single-agent summary, channel ablations & AUROC: 0.9688; median lead time: 3.0 years; no AUROC gain over best simpler baseline & Recognition dates contestable; retrospective only \\
Climate-Vector Emergence & Can cross-domain signals distinguish emergence from stable endemic activity? & 12 emergence events and 12 regional controls; first-signal years and lead-time scoring & Climate-only, ecology-only, epi-only, single-agent combined fraction & AUROC: 0.944; matched-pair acc.: 0.917; median lead: 5.0 y; perm $p<0.001$ & Curated 12+12 panel; signals literature-anchored \\
Cosmic Filter & Can four vetting channels separate confirmed planets from false positives? & 12 confirmed planets and 12 mission-era false positives; four binary literature flags plus lead time & Transit-only, shape-plus-stellar, no-archival, no-follow-up, single-agent combined fraction & AUROC: 0.955; matched-pair acc.: 1.000; median lead: 1.0 y; perm $p<0.001$ & Curated retrospective panel; dispositions may change \\
\bottomrule
\end{tabularx}
\end{table*}

\begin{table*}[h]
\centering
\small
\setlength{\tabcolsep}{5pt}
\caption{{\bf Regime map for coordinated cross-domain agents.} The benchmark portfolio separates applications where coordination improves discrimination from applications where its value is interpretive or representational.}
\label{tab:regime_map}
\begin{tabularx}{1\textwidth}{@{}>{\raggedright\arraybackslash}p{0.22\textwidth} >{\raggedright\arraybackslash}X >{\raggedright\arraybackslash}X >{\raggedright\arraybackslash}X@{}}
\toprule
Regime & Diagnostic pattern & Empirical example & Claim supported \\
\midrule
Distributed incomplete evidence & Complementary channels and lead time. & Climate-Vector Emergence; Cosmic Filter. & Coordination improves over single-channel baselines; provenance distinguishes cases tied with summary baselines. \\
Dominant single channel & One signal carries most separation. & Computational Kuhn. & Coordination adds provenance and interpretation. \\
Representational mapping & Structure recovery, not prediction. & Sound of Molecules. & Coordination exposes cross-domain structure. \\
\bottomrule
\end{tabularx}
\end{table*}

\begin{figure}[hbtp]
\centering
\includegraphics[width=1\columnwidth]{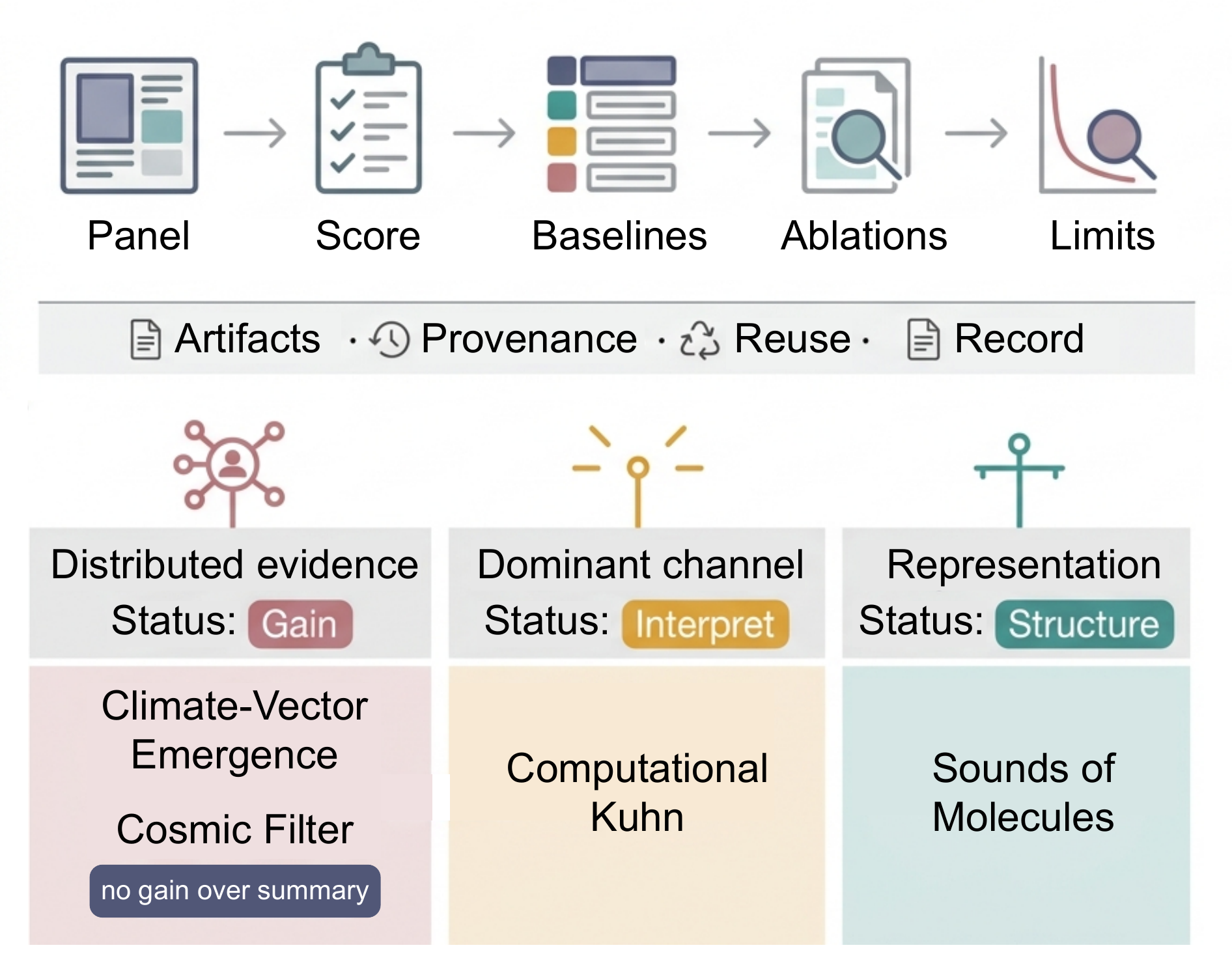}
\caption{{\bf Summary evidence map.} The benchmark design and the regime map summarize where coordination changes the supported
inference, where it mainly adds provenance and auditability, and where its
contribution is representational rather than predictive.}
\label{fig:summary_evidence_map}
\end{figure}

\subsection{A regime map identifying where coordination matters}

The cross-application comparison yields three operating regimes (Fig.~\ref{fig:summary_evidence_map}; Table~\ref{tab:regime_map}). The regime map qualifies the performance claim rather than ranking all agentic workflows on one axis. Coordination is credited only relative to explicit comparators: scripted single-channel baselines, single-agent summaries, ablations, null controls, and each benchmark's stated limitation.

The first regime is distributed incomplete evidence. Climate-Vector Emergence is the clearest example: climate suitability, ecological establishment, and epidemiological recognition arrive through different systems and at different times, so lead-time-weighted coordination improves discrimination over both single-channel baselines and a combined-fraction summary. Cosmic Filter also uses complementary channels, and its full composite improves over transit-only and shape-plus-stellar baselines, but it is effectively tied with a strong combined-summary baseline. This distinction is important: decomposition can add provenance and false-positive auditing even when the scalar score is already saturated by a summary comparator.

The second regime is dominant single-channel evidence. Computational Kuhn shows this case: citation topology already captures most retrospective separation, so semantic and funding channels add interpretive coverage and provenance without improving top-line AUROC.

The third regime is representational mapping. Sound of Molecules is not primarily a prediction benchmark; the relevant contribution is whether a chemistry-to-harmony mapping recovers interpretable class-level structure. Here coordination changes the representation and analysis trail, even though it does not outperform the strongest chemistry baseline on retrieval@3. We first examine the distributed-evidence regime in two domains, then use the remaining cases to show where coordination changes interpretation or representation rather than top-line discrimination.

\subsection{Distributed evidence in vector-borne disease emergence}

Climate-Vector Emergence tests the first distributed-evidence regime. Climate science characterizes vector suitability, ecology tracks establishment, and epidemiology records human disease, but no channel alone captures the full emergence sequence. The benchmark tests whether coordinated evidence integration across all three disciplines detects vector-borne disease emergence earlier and more reliably than scripted single-channel warning systems~\cite{lafferty2009ecology,caminade2014impact,medlock2015vectors,semenza2022climate}.

The benchmark uses a frozen matched panel of 12 documented emergence or range-expansion events, including West Nile virus in North America, Aedes albopictus in Europe, Chikungunya and Zika in the Americas, Ixodes scapularis in Canada, and Anopheles stephensi in the Horn of Africa~\cite{nash2001outbreak,medlock2015vectors,kraemer2019spread,ryan2019global,mordecai2017detecting}. Each positive case has predeclared climate, ecological, and epidemiological first-signal years and is matched against a stable endemic control in the same region. Comparator arms include scripted climate-only, ecology-only, and epidemiology-only warning models, plus a single-agent combined-signal-fraction baseline.

The full workflow achieves leave-one-pair-out AUROC 0.944, matched-pair accuracy 0.917, median positive lead time 5.0 years, and detection rate 1.00. The single-channel baselines reach AUROC 0.583 for climate-only niche expansion, 0.667 for ecology-only evidence, and 0.625 for epidemiology-only evidence. Thus the composite improves by +0.277 AUROC over the strongest single-channel baseline and by +0.208 AUROC over the single-agent combined-signal-fraction baseline (AUROC 0.736), with label-permutation \(p < 0.001\) on 2000 relabelings. The result indicates that lead-time structure contributes information beyond either isolated channels or the presence/absence of detected channels.

Comparator arms indicate that the gain is not explained by a single detected-channel flag. Climate-only, ecology-only, and epidemiology-only scoring all underperform the full composite. The supported interpretation is therefore that lead-time-weighted cross-channel scoring improves over isolated channels and the combined-fraction summary. Matched controls are central: they require the composite to discriminate emergence from stable endemic activity in the same region, rather than merely detect rising activity among positives.

The +0.277 AUROC gain over the strongest single-channel baseline, and the +0.208 gain over the combined-fraction baseline, are therefore aligned with the temporal structure that the benchmark was designed to capture.

A site-level analysis illustrates the mechanism underlying this regime. At the Dakar sentinel site, ENSO intermittency is more informative than total annual precipitation (Fig.~\ref{fig:cve_panel}). Episodic La Niña rainfall events refill artificial containers, providing a breeding-habitat pulse that precedes adult \textit{Ae.~aegypti} emergence by approximately 12 days, the aquatic-stage development lag captured by the Brière vectorial capacity curve (Fig.~\ref{fig:cve_bench}). Annual climate summaries, adult-vector surveillance, and epidemiological case reports each capture only part of this sequence. The composite links these observational windows into the pathway La Niña rainfall $\to$ container refill $\to$ 12-day aquatic development $\to$ adult emergence $\to$ transmission risk.

\begin{widefigure}[h]
\centering
\includegraphics[width=0.9\textwidth]{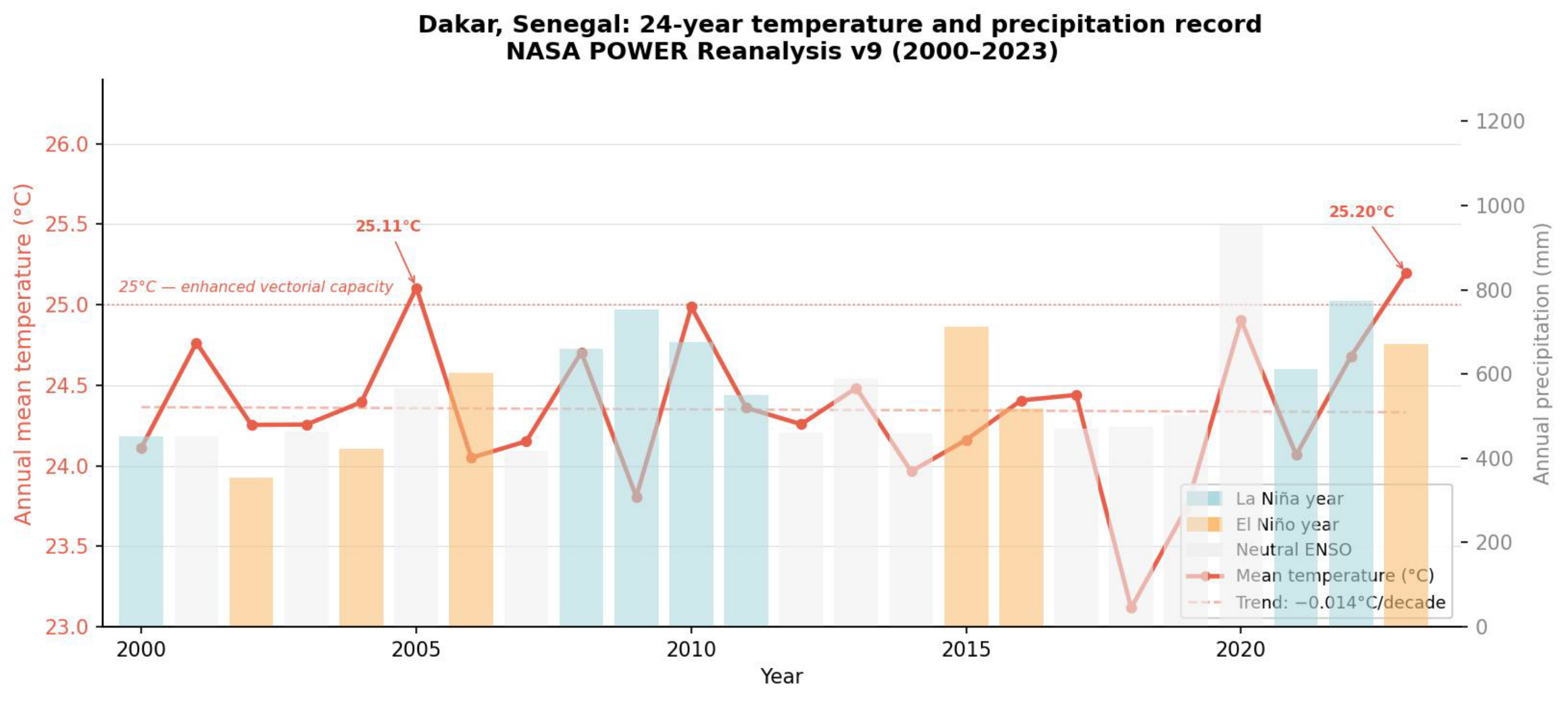}
\caption{{\bf Climate-Vector Emergence: Dakar sentinel site 24-year climate record (2000--2023).} Dual-axis plot of annual mean temperature (°C, left axis, red line with regression) and total annual precipitation (mm, right axis, bars). Precipitation bars are colored by ENSO phase (La Niña\,=\,blue, El Niño\,=\,amber, neutral\,=\,gray). Horizontal dashed lines mark the \textit{Ae.~aegypti} thermal-activity enhancement threshold (25°C) and the bioclimatic suitability floor (18°C). Temperature trend $-0.0139$°C decade$^{-1}$ (regression line); above-threshold years (2005, 2023) annotated. Data: NASA POWER Reanalysis v9.}
\label{fig:cve_panel}
\end{widefigure}

\begin{widefigure}[h]
\centering
\includegraphics[width=0.9\textwidth]{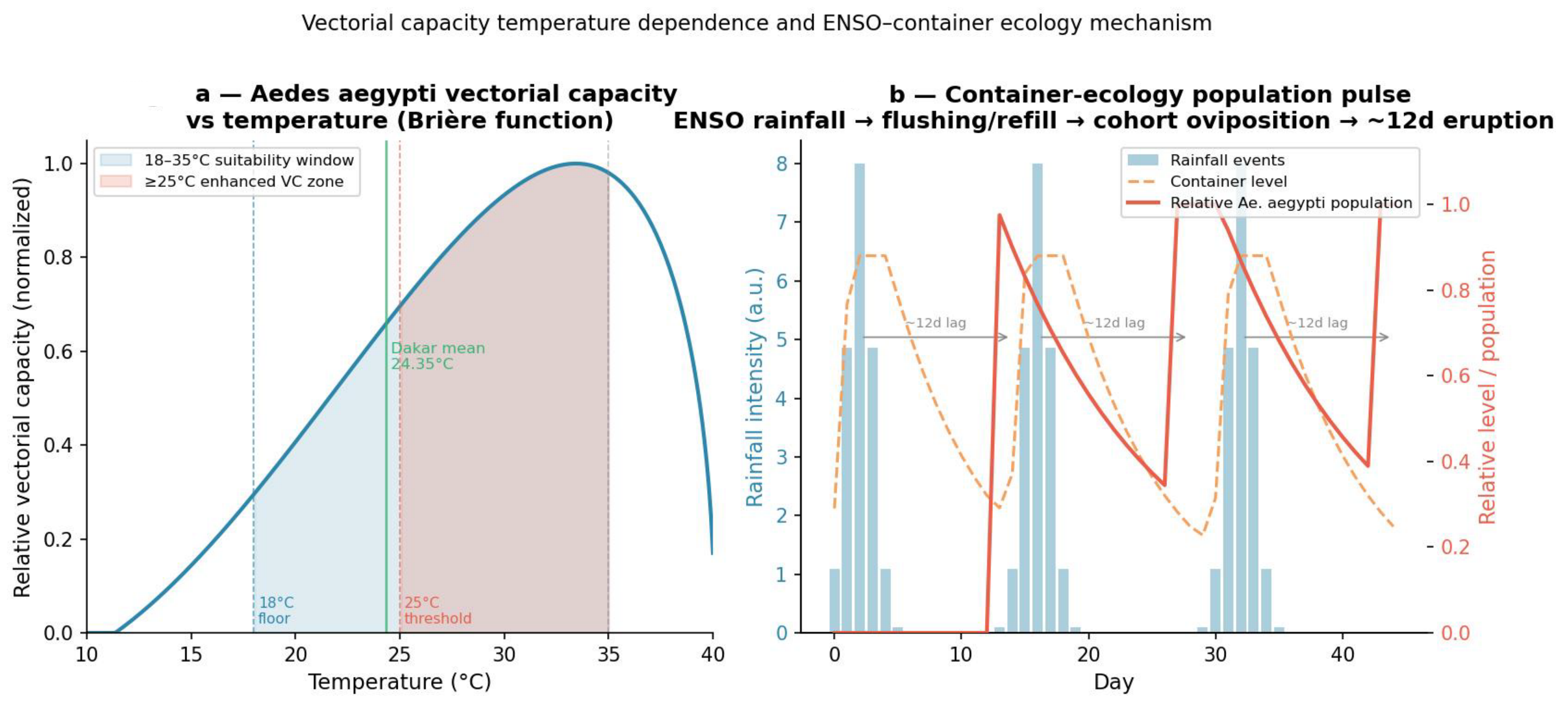}
\caption{{\bf Climate-Vector Emergence: ENSO--container ecology mechanistic pathway.} \textit{Left:} Brière vectorial capacity curve for \textit{Ae.~aegypti} (development rate vs.\ temperature; $C_T=18$°C, $T_{\max}=40.1$°C); shaded region marks the nonlinear thermal window over which vectorial capacity increases steeply. \textit{Right:} Container-habitat population-pulse schematic showing the $\sim$12-day aquatic-stage lag between rainfall-driven container refill events and adult mosquito emergence.}
\label{fig:cve_bench}
\end{widefigure}

The public-health case establishes the distributed-evidence pattern in a temporally ordered biological system. The next case asks whether the same logic holds in a very different setting: astronomical vetting, where false positives are resolved by combining independent observational records.

\subsection{A second distributed-evidence test in exoplanet vetting}

Cosmic Filter provides an independent distributed-evidence test in astronomy. Exoplanet vetting combines transit shape, stellar context, archival cross-checks, and follow-up confirmation, each addressing a distinct false-positive risk. The benchmark tests whether this four-channel composite discriminates confirmed transiting planets from matched false-positive candidates more reliably than scripted single-signal vetting pipelines~\cite{thompson2018kepler_dr25,coughlin2016robovetter,morton2016falsepositive}.

The panel comprises 12 confirmed Kepler/K2/TESS/MEarth planets matched against 12 same-era candidates later dispositioned as false positives in published vetting catalogs~\cite{quintana2014kepler186,thompson2018kepler_dr25,coughlin2016robovetter,crossfield2016k2,mayo2018k2fpp}.

The benchmark protocol is a structured literature extraction over 24 candidate rows. For each candidate, the workflow records binary evidence for planet-consistent transit geometry, uncontaminated stellar context, independent archival support, and follow-up confirmation. These flags are drawn from mission catalogs and follow-up publications rather than from re-fitting raw photometric data~\cite{thompson2018kepler_dr25,coughlin2016robovetter,morton2016falsepositive}. A vetting integrator combines the four channels with fixed weights and a small lead-time bonus, and AUROC is computed in leave-one-pair-out mode across the 12 matched pairs.

The full composite achieves leave-one-pair-out AUROC 0.955, matched-pair accuracy 1.000, and median positive lead time 1.0 years. The transit-shape-only baseline reaches AUROC 0.708 and shape-plus-stellar reaches 0.781; the full composite exceeds both, with label-permutation $p<0.001$. A single-agent combined-fraction baseline reaches AUROC 0.951, effectively tied with the full composite. Thus, multi-agent decomposition does not improve AUROC over a strong combined-summary baseline here; its contribution is channel-specific provenance and matched false-positive auditing.

\begin{figure}[hbtp]
\centering
\includegraphics[width=1.\columnwidth]{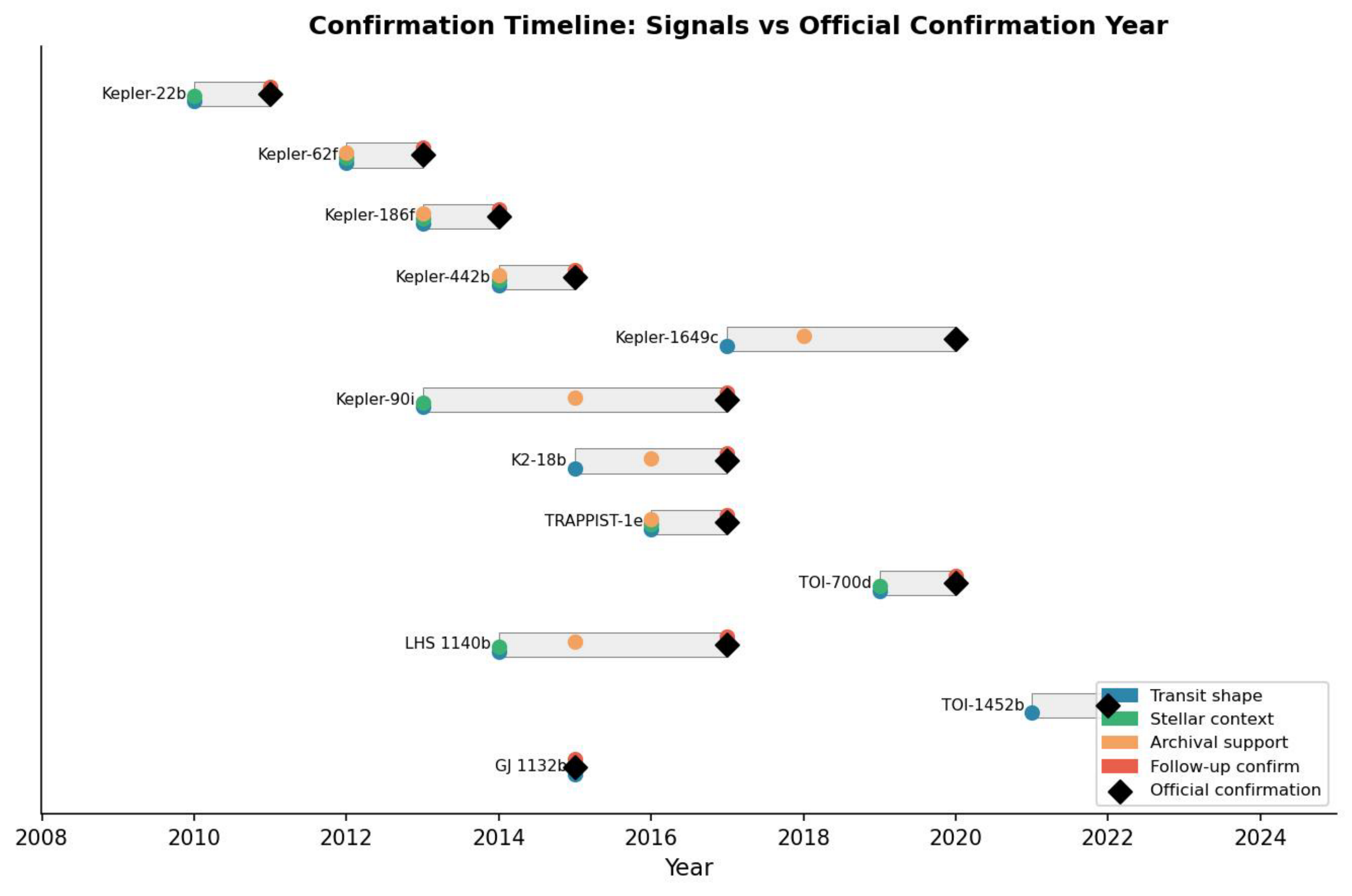}
\caption{{\bf Cosmic Filter: per-channel lead time to positive detection.} For each of the 12 confirmed planets, the plot shows the number of years elapsed between the initial positive signal in each vetting channel and the community-recognized confirmation date. Transit shape and stellar context typically occur earliest, often at initial survey release; archival cross-checks and follow-up confirmation accumulate later as independent datasets become available.}
\label{fig:cf_panel}
\end{figure}

\begin{figure}[hbtp]
\centering
\includegraphics[width=1.\columnwidth]{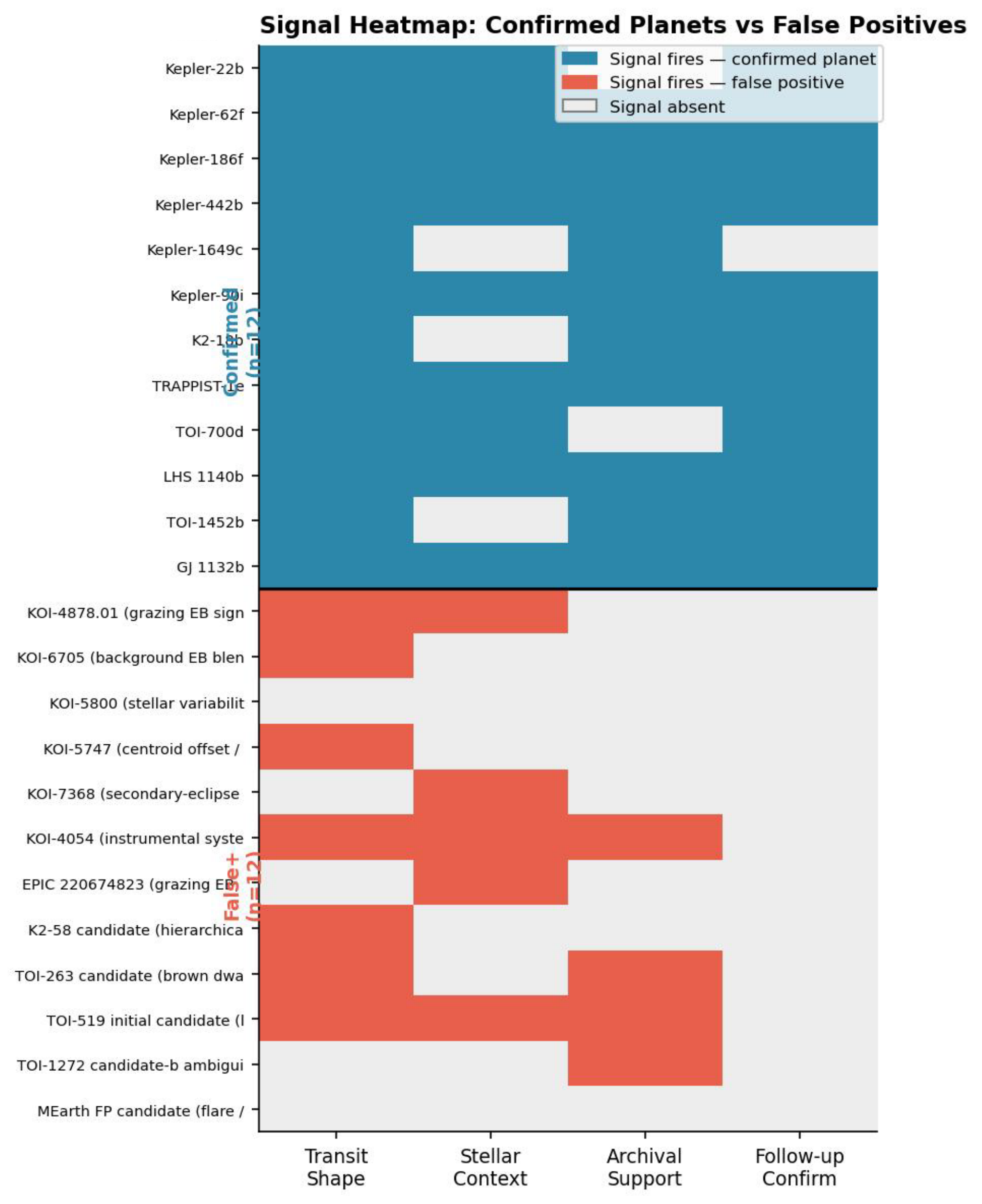}
\caption{{\bf Cosmic Filter: per-candidate signal presence matrix.} Binary signal flags (0 = absent, 1 = present) for all 24 candidates across four vetting channels. Confirmed planets (top block) accumulate positive flags across channels; false positives (bottom block) show gaps, most prominently in follow-up confirmation, the strongest removal-test channel. The matrix summarizes the cross-channel complementarity used by the composite score.}
\label{fig:cf_bench}
\end{figure}

Ablations identify follow-up confirmation as the strongest removal test: dropping it reduces AUROC from 0.955 to 0.851. The remaining partial-channel comparisons show that one- or two-channel evidence is insufficient to match the full composite, while the near-tie with the combined-fraction baseline indicates that decomposition mainly contributes channel-specific provenance and false-positive auditing in this panel. The channel lead-time structure and per-candidate signal matrix are shown in Figs.~\ref{fig:cf_panel} and~\ref{fig:cf_bench}.

Together, Climate-Vector Emergence and Cosmic Filter define the distributed-evidence regime, but with different strengths of claim. Both use frozen matched retrospective panels of 12 positives and 12 controls, evaluate composite signals against scripted single-channel baselines, and achieve AUROC above 0.94 with permutation $p<0.001$. Climate-Vector Emergence also exceeds the combined-fraction summary baseline, supporting a scalar discrimination gain from lead-time-weighted evidence integration. Cosmic Filter is effectively tied with its combined-fraction summary baseline, so its supported contribution is decomposition, provenance, and matched false-positive auditing rather than a unique AUROC gain over summary scoring. The next case provides a contrast: a setting where one channel already carries most of the discriminative signal.

\subsection{Dominant-channel evidence in paradigm-shift discrimination}

Computational Kuhn evaluates the dominant-channel regime, where coordination is useful but not primarily because it improves discrimination. The benchmark tests whether composite signals from citation topology, semantic drift, and funding flow improve retrospective discrimination of scientific paradigm shifts relative to simpler bibliometric workflows~\cite{k1962}. The application is cross-domain because the shift panel spans disjoint scientific fields, including geology (plate tectonics, $\sim$1965), molecular biology (CRISPR-Cas9, $\sim$2012), genomics (single-cell RNA sequencing, $\sim$2018), and neuroscience (optogenetics, $\sim$2010), and because the composite combines three data modalities with no natural common scale.

The benchmark is restricted to a frozen summary panel of 16 historical shifts and 16 matched non-shift controls, with recognition dates predeclared before scoring. Each example is scored from summary-level citation, semantic, and funding features. The comparator arms are an equal-weight single-agent summary, a scripted bibliometrics baseline, and single-channel ablations. This design supports retrospective discrimination, not prospective forecasting from raw local corpora. The citation-growth panel and composite early-warning trace are summarized in Figs.~\ref{fig:ck_panel} and~\ref{fig:ck_ablation}.

The full workflow achieves leave-one-pair-out AUROC 0.9688, pairwise accuracy 0.9375, and median lead time 3.0 years. Simpler baselines match this retrospective performance: both the equal-weight single-agent summary and the scripted bibliometrics baseline achieve AUROC 0.969. The coordinated workflow therefore does not provide a measured AUROC advantage in this application. Its contribution is the integration of heterogeneous historical signals within an auditable analysis record. Calibration remains imperfect (calibration error 0.6243), and the scores should not be interpreted as prospective probabilities. Ablations identify citation topology as the dominant channel: citation-only scoring reaches AUROC 0.969, whereas semantic drift falls to 0.844 and funding is near chance at 0.531.

\begin{widefigure}[h]
\centering
\includegraphics[width=0.95\textwidth]{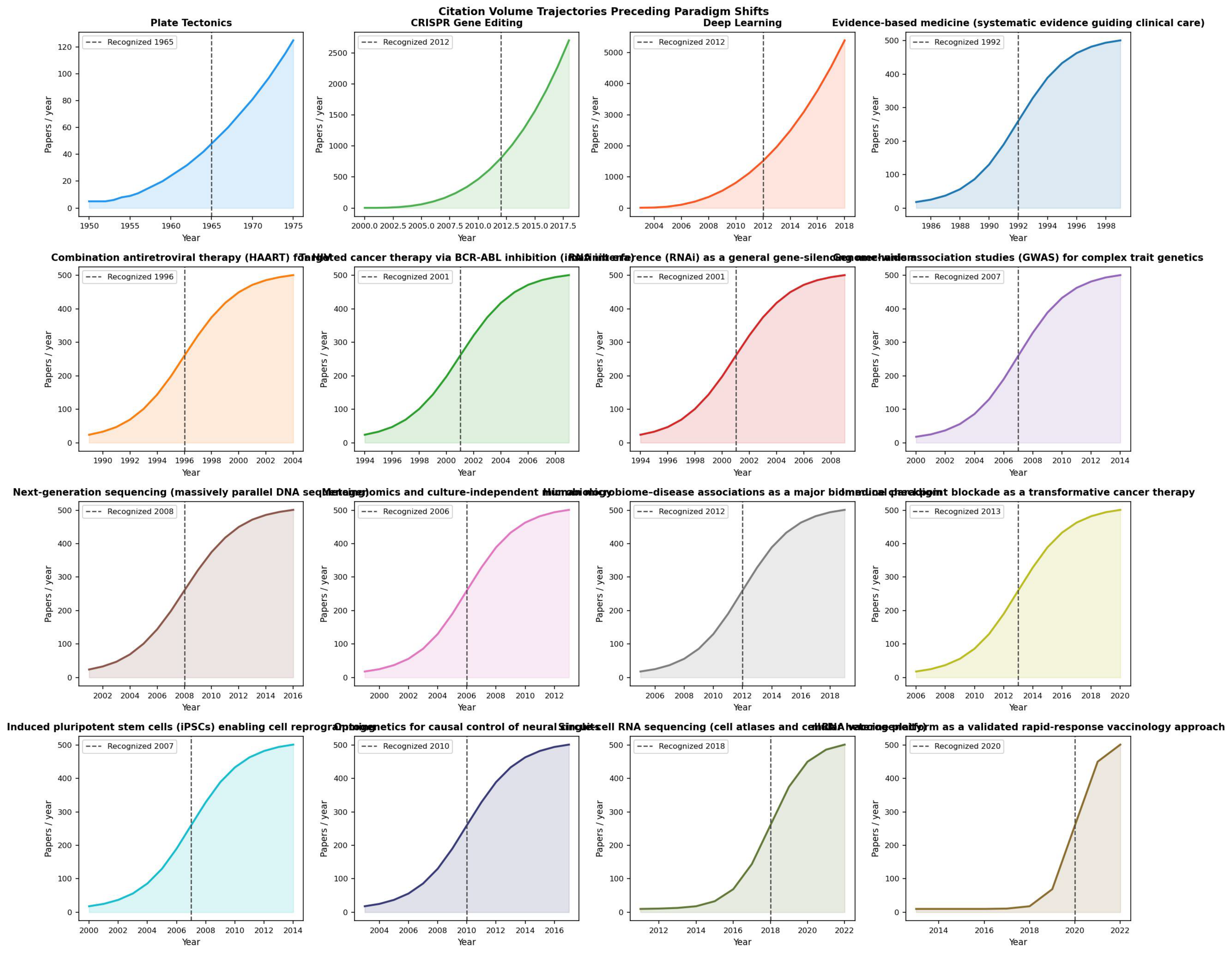}
\caption{{\bf Computational Kuhn: citation growth curves across 16 paradigm shifts.} Normalized annual publication counts for each of 16 historical paradigm shifts (colored curves), aligned to the predeclared recognition year ($t=0$, orange dotted line). The dashed white curve shows the mean S-curve trajectory. The shaded region marks the 2--3 years before $t=0$ used to summarize pre-recognition acceleration. The 16 shifts span multiple disciplines (geology, genomics, biomedicine, neuroscience).}
\label{fig:ck_panel}
\end{widefigure}

\begin{widefigure}[hbtp]
\centering
\includegraphics[width=.95\textwidth]{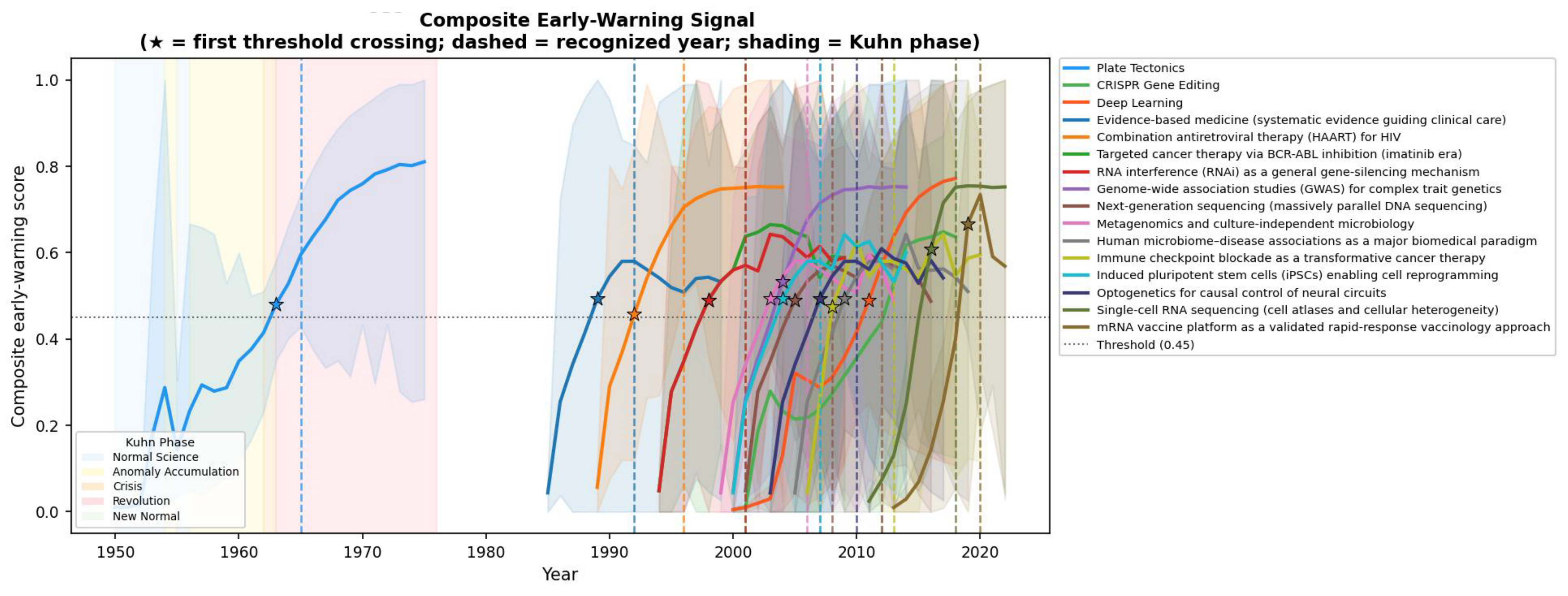}
\caption{{\bf Computational Kuhn: composite early-warning signal with Kuhn phase annotations.} Four-panel stacked plot showing citation acceleration (blue), semantic drift (green), and funding intensity (purple) channels, plus their weighted composite (orange), over the ten years preceding paradigm recognition ($t=0$, white dashed). Background shading marks Kuhn phases: pre-paradigmatic, anomaly accumulation, crisis, and revolution. The marked threshold crossing occurs at median $t=-3$\,years across the 16-shift panel, providing the lead-time estimate. Citation topology alone dominates the composite (AUROC 0.969 vs.\ 0.844 for semantic and 0.531 for funding in isolation); the composite adds interpretive coverage but not a measured AUROC advantage over the citation channel alone.}
\label{fig:ck_ablation}
\end{widefigure}

This case shows one limit of coordination: when a strong single channel already separates positives from controls, additional channels mainly improve interpretation and auditability. The final case shifts the question further, from discrimination to whether coordination can construct a useful cross-domain representation.

\subsection{Representational mapping in molecular sonification}

Sound of Molecules tests this representational regime. Rather than asking whether coordination improves classification, the benchmark evaluates whether a deterministic mapping from molecular descriptors to harmonic representations recovers interpretable chemical organization rather than an arbitrary aesthetic correspondence. This application follows earlier work linking molecular mechanics, natural language, and musical representations through attention-based models~\cite{buehler2022attention}. The scientific claim is limited to cross-domain structure recovery: molecules that are topologically distant in chemical fingerprint space may become proximal in harmonic space, and non-random proximity with respect to pharmacological class provides evidence for structure not captured by the source-domain metric.

\begin{figure}[h]
\centering
\includegraphics[width=1.\columnwidth]{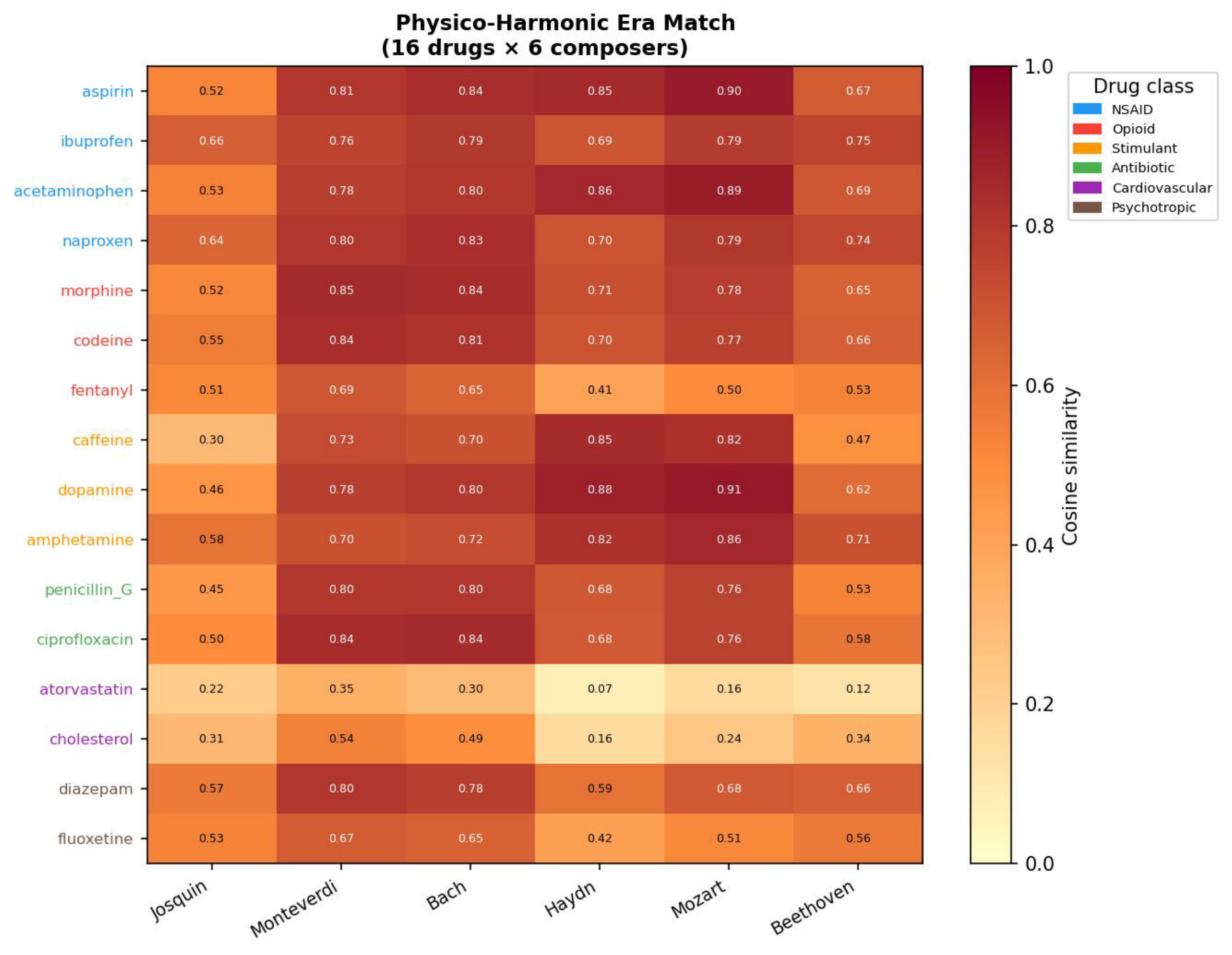}
\caption{{\bf Sound of Molecules: era-match heatmap.} Pairwise similarity scores between 16 drug compounds spanning six pharmacological classes and 6 composers spanning the Baroque through Modern eras. Scores are derived from RDKit descriptor vectors projected onto harmonic-feature embeddings; color scale runs from low (dark) to high (pale blue). Block structure is evaluated against shuffled-label and random-mapping controls ($p=0.0178$).}
\label{fig:som_main}
\end{figure}

The benchmark uses a fixed 16-compound panel spanning NSAIDs, opioids, stimulants, antibiotics, cardiovascular agents, and psychotropics. Each compound is embedded through a sonification pipeline built from physicochemical descriptors~\cite{rdkit} and music21 composer corpus statistics~\cite{music21}. The comparator arms are Morgan fingerprints, physicochemical cosine similarity, shuffled-label nulls, random descriptor-to-harmony mappings, and a no-3D ablation. The primary readouts are retrieval@3, same-class nearest-neighbor coherence, and robustness across descriptor variants; the resulting compound--composer heatmap and physicochemical embedding are shown in Figs.~\ref{fig:som_main} and~\ref{fig:som_bench}.

The full workflow achieves retrieval@3 of 0.2708, below the best scripted chemistry baseline (0.2917), but its same-class nearest-neighbor rate reaches 0.6875, against 0.3125 for Morgan fingerprints and 0.1875 for the physchem cosine baseline. The embedding is 7.2\% worse on retrieval@3 and 120.0\% better on same-class nearest-neighbor coherence than the best relevant baseline.

The controls support this restricted interpretation. Retrieval@3 under the observed mapping exceeds the shuffled-compound-label null mean of 0.125 ($p=0.0178$), and the nearest-neighbor rate exceeds random expectation ($p=0.0004$). A random descriptor-to-harmony control yields null mean retrieval@3 of 0.224, below the observed mapping. The no-3D sonic ablation lowers retrieval@3 to 0.250 and nearest-neighbor coherence to 0.375. Across descriptor variants, the robustness score is 0.8021; clustering diagnostics (ARI 0.083, NMI 0.557) and scaffold-held-out accuracy of 0.500 provide additional conservative evidence of non-random structure. Several same-class pairs that are weak under 2D fingerprints become close in the sonic representation, including diazepam and fluoxetine in the psychotropic class. The result supports a structure-recovery claim, not a claim of superior retrieval performance.

\begin{widefigure}[h]
\centering
\includegraphics[width=0.8\textwidth]{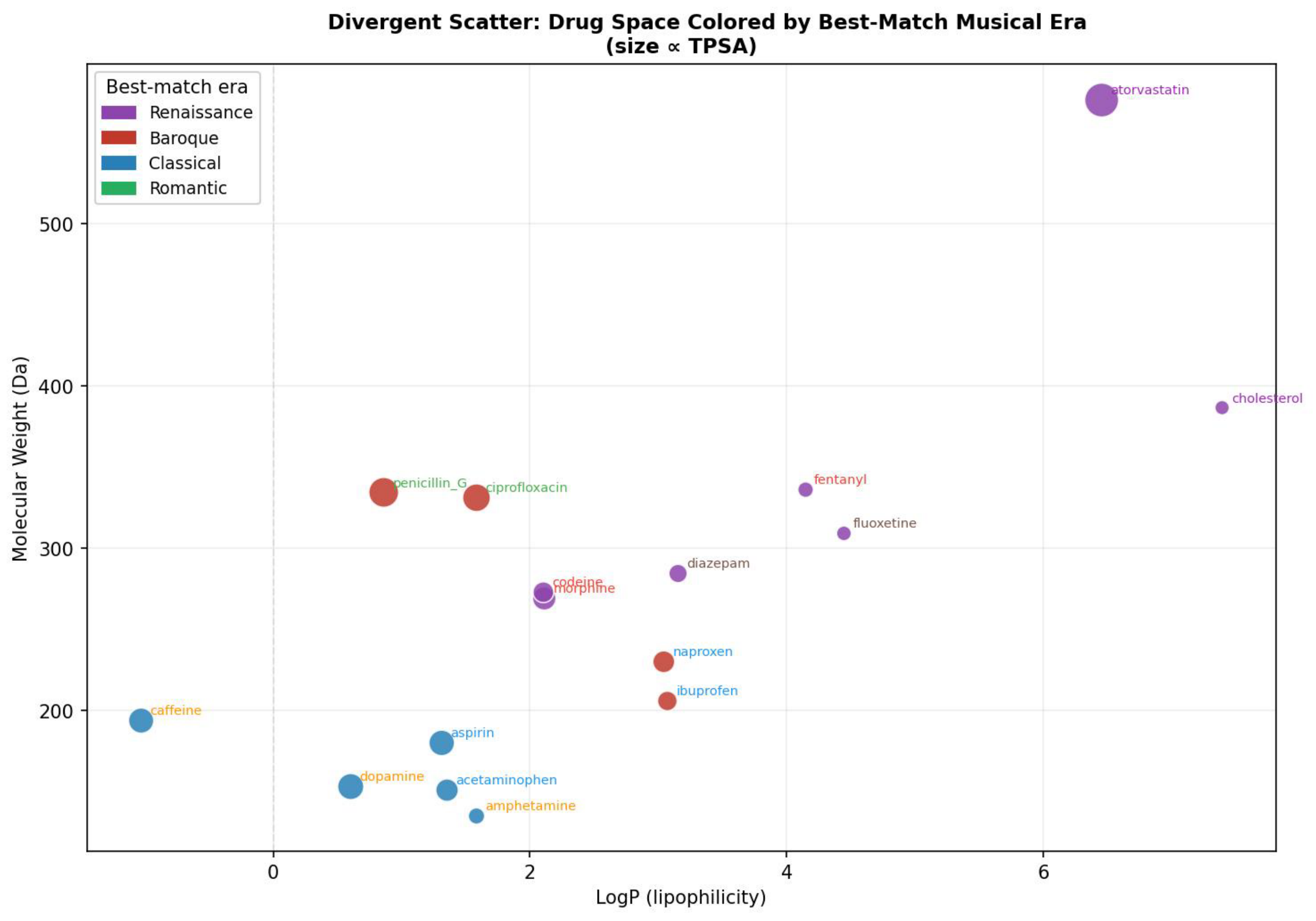}
\caption{{\bf Sound of Molecules: physicochemical space colored by era assignment.} Scatter of 16 drug compounds in LogP--molecular-weight space; each point is colored by its dominant musical-era assignment (Baroque = violet, Classical = blue, Romantic = red, Modern = green). Lipinski drug-likeness boundaries (MW\,=\,500\,g\,mol$^{-1}$, LogP\,=\,5) are shown as dashed lines.}
\label{fig:som_bench}
\end{widefigure}

Across the dominant-channel and representational regimes, coordination therefore adds value in different ways: by preserving an interpretable multi-signal account when performance is already saturated, or by producing an auditable mapping whose value is structural rather than predictive. The next subsection describes the artifact layer that makes those claims inspectable across all four applications.

\subsection{Artifact preservation and public investigation records}

The benchmark and regime map require more than final scores; they require records that show how each result was assembled, what intermediate evidence was reused, and where simpler comparators remain sufficient. ScienceClaw~$\times$~Infinite provides this supporting layer through artifact preservation, provenance, reuse, public investigation records, and continuation after partial results~\cite{wang2026scienceclaw_infinite}.

This record layer keeps the four applications from functioning as isolated demonstrations. The relevant unit of computation is not an isolated model response but a provenance-preserving sequence of intermediate artifacts that can be consumed by subsequent analytical stages (Fig.~\ref{fig:engine}). In the four applications, those artifacts support descriptor-to-harmony transfer for molecular structure recovery, multimodal retrospective scoring for paradigm-shift discrimination, climate--ecology--epidemiology fusion for emergence detection, and four-channel literature-based vetting for exoplanet candidates.

\begin{widefigure}[hbtp]
\centering
\includegraphics[width=0.95\textwidth]{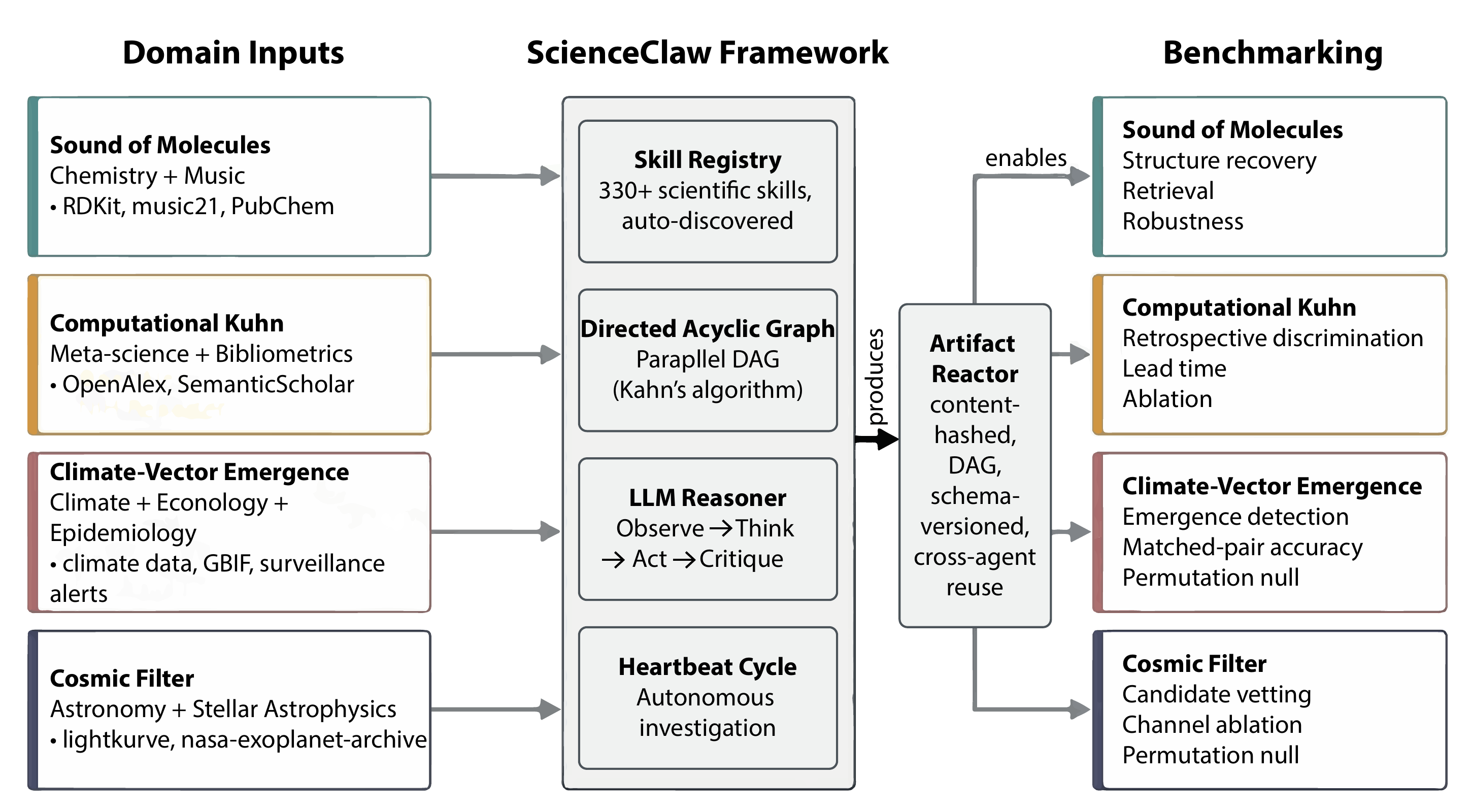}
\caption{{\bf Artifact-mediated benchmark infrastructure.} Domain inputs for the four applications (left) are routed through the ScienceClaw~$\times$~Infinite workflow, including the Skill Registry, DAG, LLM Reasoner, and Heartbeat Cycle. The engine produces a content-hashed, schema-versioned artifact store with provenance DAGs and cross-agent reuse (center). Stored artifacts then support validation and benchmarking panels (right) for molecular structure recovery, retrospective paradigm-shift discrimination, climate--vector emergence detection, and transiting-exoplanet vetting.}
\label{fig:engine}
\end{widefigure}

Together, these results support a structured benchmark account of where coordination matters. Frozen panels, explicit baselines, ablations, nulls, and limitations make coordinated-agent results comparable across domains. The cross-application comparison yields a regime map rather than a single ranking of agentic workflows: coordination can improve measured discrimination when distributed evidence and lead-time structure add signal, can mainly add provenance when summary or single-channel baselines already saturate AUROC, and can support representation-level structure recovery when prediction is not the primary endpoint. The platform contribution is the auditable infrastructure underlying this evaluation.

\section{Discussion}
This paper treats cross-domain integration as a scientific object: when does combining partial evidence across disciplines change the result, and when does it change how the result is interpreted? The four applications support the regime map summarized in Fig.~\ref{fig:summary_evidence_map} and Table~\ref{tab:regime_map}. In Climate-Vector Emergence, evidence is genuinely distributed across climate, ecology, and epidemiology, and the coordinated composite exceeds both scripted single-channel baselines and the single-agent combined-fraction baseline. In Cosmic Filter, the four-channel composite clearly improves over transit-shape and shape-plus-stellar baselines, but is effectively tied with a strong single-agent combined-fraction summary. In Computational Kuhn, citation topology already captures most discriminatory power, so multi-channel aggregation adds interpretive coverage and provenance rather than top-line AUROC. In Sound of Molecules, the scientific value is representational: a harmony embedding recovers class-level neighborhood structure not captured by source-domain baselines, despite not outperforming the strongest chemistry baseline on retrieval@3.

The evidence also defines the limits of the claim. Multi-agent decomposition does not automatically outperform a strong single-agent combined baseline. Cosmic Filter is effectively tied with the combined-fraction single-agent baseline, and Computational Kuhn is matched by both the scripted bibliometric baseline and the equal-weight summary baseline. These results argue against treating agent count or decomposition as an intrinsic source of performance. Coordinated agents are most valuable when evidence is distributed across incomplete disciplinary channels, when temporal ordering or channel-specific provenance changes the supported inference, and when auditability and reuse of intermediate artifacts matter for the scientific claim.

The applications clarify this boundary. Climate-Vector Emergence benefits from ordered signals across climate, ecology, and epidemiology. Cosmic Filter benefits from separating transit, stellar, archival, and follow-up evidence, even though a strong combined summary performs similarly. Computational Kuhn shows the opposite case: when citation topology dominates, coordination mainly supports interpretation. Sound of Molecules defines a third case, where the useful output is a cross-domain representation rather than a better predictor.

The computing contribution is therefore a benchmarkable workflow pattern: preserve intermediate artifacts, make provenance inspectable, transfer evidence across disciplinary tools, and evaluate the resulting synthesis against frozen panels and explicit comparators. Table~\ref{tab:regime_map} ties each operating regime to its diagnostic pattern, empirical example, and supported claim. Limitations remain as the portfolio is modest in scale, the panels are curated and retrospective, and deterministic pipelines support reproducibility but do not test prospective performance.

\section{Methods}

\subsection{Artifact-mediated scientific workflows}

The computing contribution of this work is the use of artifact-mediated scientific workflows as the unit of reproducible analysis. Each specialist agent produced intermediate products as structured artifacts rather than only free-text messages. An artifact was represented as
\[
a=(u,\; y,\; m,\; P,\; q,\; r),
\]
where $u$ is a unique artifact identifier, $y$ is the typed payload, $m$ is metadata including producer identity, tool name, timestamp, application identifier, and schema version, $P$ is the set of parent artifact identifiers, $q$ is a result-quality flag, and $r$ is a short human-readable summary. Payloads included manifests, feature tables, literature-extraction records, score tables, figures, and narrative syntheses. To make artifacts reusable and auditable, each payload was serialized as canonical JSON and assigned a content address
\[
h(a)=\mathrm{SHA256}(\mathrm{canonical\_JSON}(a)).
\]
Derived artifacts retained the identifiers of their parents, producing a provenance graph $G=(V,E)$, where each node is an artifact and each directed edge links a derived artifact to an upstream artifact it consumed. Reuse occurred by passing content-addressed artifacts, not by copying free-text summaries between agents. This made cross-domain evidence transfer explicit: a downstream synthesis could consume, for example, a molecular descriptor table, a composer-profile table, or a disease first-signal table only through its artifact identifier and hash. Public investigation records preserved the narrative synthesis together with artifact summaries and hashes, allowing later readers to trace a conclusion back to the intermediate products that supported it. The main benchmark artifacts were frozen before manuscript drafting, and reported tables were generated from those frozen artifacts rather than manually re-entered. Additional benchmark-arm and control tables are provided in the Supplementary Information.

\subsection{Common benchmark design}

Each application used the same benchmark structure: a full coordinated workflow, scripted single-signal baselines, a strong single-agent summary baseline when the task admitted one, ablations that removed channels or feature classes, null controls, primary metrics, and an explicit limitation statement. A benchmark panel is the fixed set of positive and control examples scored in an application. The full workflow is the composite score produced after cross-domain evidence transfer and synthesis. Scripted baselines use one prespecified evidence channel without coordination. Single-agent baselines summarize the same available signals without specialist decomposition. Panels, control definitions, comparator definitions, scoring code, and statistical tests were fixed before scoring; when weights or thresholds were selected, selection was performed within the declared training fold and never used held-out labels.

For examples indexed by $i$ and evidence channels indexed by $j$, the general composite score was
\[
S_i=\sum_j w_j x_{ij}+\sum_j \gamma_j \ell_{ij},
\]
where $x_{ij}$ is a binary or normalized channel feature, $w_j$ is its feature weight, $\ell_{ij}$ is normalized lead time, and $\gamma_j$ is its lead-time weight. Lead time was defined as
\[
\ell_{ij}=\frac{\max(0,T_i-t_{ij})}{W_i},
\]
where $T_i$ is the recognition, confirmation, or disposition year; $t_{ij}$ is the first year in which channel $j$ supplied evidence; and $W_i$ is the evaluation-window length. For matched panels, matched-pair accuracy was
\[
\frac{1}{n}\sum_{i=1}^{n}\mathbf{1}[S_i^+>S_i^-],
\]
where $S_i^+$ and $S_i^-$ are the positive and matched-control scores. AUROC was computed from the pooled positive and control scores. For the matched positive/control panels, label-permutation tests randomly reassigned positive/control labels 2000 times and compared the observed AUROC with the resulting null distribution; the molecular-sonification null tests used 5000 permutations as described below.

\subsection{Workflow algorithm}

Each application followed the same computational algorithm. First, the authors froze the application panel $B=\{(i,y_i,c_i)\}$, where $i$ is the item identifier, $y_i$ is the positive/control or class label, and $c_i$ contains application-specific metadata such as recognition year, confirmation year, pharmacological class, or matched-control identity. Second, specialist agents collected domain evidence with designated tools and stored each intermediate product as a content-addressed artifact. Third, fixed feature transforms converted artifacts into feature matrices $X\in\mathbb{R}^{n\times m}$ and, where applicable, lead-time matrices $L\in\mathbb{R}^{n\times m}$. Fourth, the full workflow and every comparator arm were scored from the same frozen $X$ and $L$ using fixed weights or declared fold-internal weight and threshold selection rules. Fifth, the evaluation layer computed AUROC, matched-pair accuracy, retrieval, nearest-neighbor coherence, confidence intervals, and permutation nulls as appropriate for the application. Sixth, the final synthesis artifact was generated from the frozen metric artifacts and linked back to the artifact graph.

In notation, a reproducible application result is
\[
R=\Phi(B,\mathcal{A}^{\ast},\theta),
\]
where $\mathcal{A}^{\ast}$ is the frozen artifact set, $\theta$ is the declared set of weights, thresholds, fold-internal selection rules, and comparator definitions, and $\Phi$ is the scoring and reporting function. This separation is important: LLM calls could assist with evidence search, tool routing, extraction checks, and prose synthesis, but benchmark numbers were computed by scripted scoring functions applied to frozen artifacts.

\begin{table*}[hbtp]
\centering
\small
\caption{{\bf LLM and tool use by application.} LLM-mediated agent steps supported evidence navigation, extraction checks, and synthesis. Structured tools and scripted scoring functions produced the benchmark metrics.}
\label{tab:methods_llm_tools}
\setlength{\tabcolsep}{4pt}
\begin{tabularx}{\textwidth}{@{}>{\raggedright\arraybackslash}p{0.20\textwidth} >{\raggedright\arraybackslash}p{0.25\textwidth} >{\raggedright\arraybackslash}X >{\raggedright\arraybackslash}X@{}}
\toprule
Application & LLM-mediated agent role & Tools and data sources & Frozen outputs \\
\midrule
Molecular sonification & Chemistry/music interpretation, tool routing, artifact summaries, and synthesis checks. & RDKit molecular descriptors and conformers; music21 chord and Roman-numeral extraction; numpy/scikit-learn similarity, retrieval, bootstrap, and permutation routines. & Compound feature matrices, composer-profile matrices, harmonic embeddings, retrieval scores, nearest-neighbor scores, null controls. \\
Retrospective paradigm-shift detection & Literature triage, extraction checks, and synthesis across citation, semantic, and funding evidence. & OpenAlex, PubMed, arXiv, bibliographic search, text-mining utilities, network-analysis tools, and fixed summary-feature scoring. & Recognition-year panel, matched controls, summary feature matrix, leave-one-pair-out AUROC, lead-time and ablation tables. \\
Vector-borne disease emergence early warning & Climate, ecology, and epidemiology evidence extraction and cross-domain synthesis. & Cited climate/vector/epidemiology literature, public surveillance descriptions, NASA POWER summaries for the Dakar analysis, and fixed lead-time scoring. & First-signal-year table, climate/ecology/epidemiology channel flags, matched-pair scores, permutation-null results. \\
Transiting-exoplanet vetting & Mission, catalog, archival, stellar-context, and follow-up evidence organization. & Published Kepler, K2, TESS, MEarth, and false-positive-vetting sources; public astronomy catalog cross-checks; fixed binary-channel and lead-time scoring. & Candidate signal matrix, confirmation/disposition anchors, composite and comparator scores, permutation-null results. \\
\bottomrule
\end{tabularx}
\end{table*}

\subsection{Molecular sonification workflow}

The molecular sonification application used a fixed 16-compound manifest spanning six pharmacological classes. Molecular structures were represented as SMILES strings and featurized with RDKit~\cite{rdkit}. The descriptor set included molecular weight, octanol-water partition coefficient, topological polar surface area, rotatable-bond count, hydrogen-bond acceptors and donors, fraction of sp\(^3\) carbons, ring counts, aromatic-ring counts, heavy-atom count, Bertz complexity, and molar refractivity. Three-dimensional features were computed from fixed-seed ETKDG conformers when embedding succeeded, followed by MMFF or UFF optimization; these features included radius of gyration, asphericity, eccentricity, inertial shape factor, normalized principal moments, spherocity, and a binned interatomic-distance histogram. The no-3D ablation used only two-dimensional descriptors.

Composer profiles were generated with music21~\cite{music21}. For each composer corpus, scores were chordified, keys were estimated, chords were mapped to Roman numerals, and chord-frequency vectors were computed over the first 64 chords per piece. The executable workflow had four stages: compound featurization, composer-profile construction, descriptor-to-harmony projection, and validation against chemistry baselines. Molecular descriptor vectors were z-scored and mapped with a fixed transform to harmonic channels representing chromatic complexity, tonal polarity, contrapuntal density, voice-leading clarity, and related harmonic features. Embeddings were L2-normalized, and pairwise similarities were computed by cosine similarity. The main chemistry baselines were Morgan radius-2 fingerprints with 2048 bits scored by Tanimoto similarity, and cosine similarity over standardized physicochemical descriptors. Null controls shuffled compound labels, shuffled composer labels, or replaced the fixed descriptor-to-harmony transform with random linear mappings. LLM-assisted agents were used to coordinate the chemistry/music interpretation and to draft artifact summaries; RDKit, music21, and numerical routines produced the feature values and scores.

Drug-class retrieval at rank $k$ was
\[
\mathrm{retrieval@}k=\frac{1}{N}\sum_i\frac{1}{k}\sum_{r\in\mathcal{N}_k(i)}\mathbf{1}[y_r=y_i],
\]
where $\mathcal{N}_k(i)$ is the set of the $k$ nearest neighbors of molecule $i$. Same-class nearest-neighbor coherence was the fraction of compounds whose top nearest neighbor shared the same pharmacological class. Bootstrap confidence intervals used 2000 resamples, and label-permutation tests used 5000 permutations for the sonification metrics.

\subsection{Retrospective paradigm-shift detection}

The paradigm-shift application used a frozen retrospective panel of 16 recognized historical shifts and 16 matched non-shift controls. Each positive example had a predeclared recognition year and an evaluation window; its matched control inherited the same window. Specialist agents assembled citation-topology, semantic-drift, and funding-flow evidence using bibliographic and literature-search tools, including OpenAlex, PubMed, arXiv, network-analysis tools, statistical modeling tools, and text-mining utilities. Because raw corpus coverage was uneven across historical periods, the benchmark was restricted to frozen summary features rather than presented as a fully prospective raw-corpus forecasting system.

For each example, the summary feature vector was
\[
\mathbf{x}_i=(d_i,\; L_i/W_i,\; L_i/5,\; 1-p_i,\; q_i),
\]
where $d_i$ is a detected-signal flag, $L_i$ is positive lead time in years, $W_i$ is the evaluation-window length, $p_i$ is the signal significance value recorded in the frozen panel, and $q_i$ is the incommensurability or vocabulary-divergence feature. In the full leave-one-pair-out composite evaluation, one positive/control pair was held out, weights and a threshold were selected using only the remaining pairs, and the held-out pair was scored without using its label information for model selection. Comparator arms included an equal-weight single-agent summary, a lead-time-only bibliometric baseline, a detection-only baseline, a significance/incommensurability baseline, and an incommensurability-only ablation. Primary readouts were AUROC, pairwise accuracy, median lead time, precision at a fixed threshold, calibration error, and ablation behavior.

The computational workflow had five stages: freeze positive and matched-control topics; retrieve bibliographic and textual evidence; extract citation-growth, vocabulary-divergence, and funding-flow summaries; store extracted evidence as artifacts with provenance links to source records; and score the frozen summary matrix. LLM-assisted agents were used for literature triage, extraction sanity checks, and synthesis of the historical narrative. They did not choose recognition years after scoring and did not directly assign the benchmark labels.

\subsection{Vector-borne disease emergence early-warning}

The vector-borne disease application used a frozen matched retrospective panel of 12 documented emergence or range-expansion events and 12 stable endemic controls in the same regions~\cite{lafferty2009ecology,caminade2014impact,medlock2015vectors,ryan2019global,mordecai2017detecting,nash2001outbreak,reisen2003epidemiology,paz2015climate,kraemer2019spread}. Each positive event had predeclared climate, ecological, and epidemiological first-signal years. Control examples were selected from related vectors or pathogens in the same region and period; they could carry nonzero climate or ecological exposure signals, but lacked a comparable epidemiological emergence signal.

The evidence channels were climate-niche drift, ecological establishment, and epidemiological surveillance. Climate evidence used climate anomaly and suitability concepts from the cited literature; ecological evidence used vector-establishment, range, habitat, or occurrence signals; and epidemiological evidence used autochthonous cases, sentinel surveillance, outbreak reports, or genomic/phylogeographic evidence. The full score used fixed weights of 0.20 for each channel-detection flag, 0.15 for climate lead time, 0.15 for ecological lead time, and 0.10 for epidemiological lead time. The strong single-agent baseline used only the combined detected-channel fraction. Scripted baselines used climate only, ecology only, or epidemiology only. The primary readouts were leave-one-pair-out AUROC, matched-pair accuracy, median positive lead time, event detection rate, and a 2000-draw label-permutation null.

The computational workflow had four stages: curate matched emergence and stable-endemic examples; extract first-signal years for climate, ecology, and epidemiology; normalize lead times within the evaluation window; and score the full composite and single-channel baselines. Climate evidence used public climate and reanalysis-derived summaries where available, including the NASA POWER record used for the Dakar sentinel-site analysis. Ecological and epidemiological evidence was extracted from the cited literature and public surveillance descriptions. Row-level source keys for the event and control manifests are retained in the frozen benchmark artifact. LLM-assisted agents supported literature search, evidence extraction, and cross-domain synthesis, while the channel flags, lead times, weights, AUROC, matched-pair accuracy, and permutation tests were frozen artifacts.

\subsection{Transiting-exoplanet vetting}

The exoplanet-vetting application used a frozen matched retrospective panel of 12 confirmed transiting exoplanets and 12 same-era candidates dispositioned as false positives in the published vetting literature~\cite{thompson2018kepler_dr25,coughlin2016robovetter,morton2016falsepositive,crossfield2016k2,mayo2018k2fpp,batalha2013planetary}. Specialist agents organized evidence into four binary literature-documented channels: transit-shape consistency, stellar-context consistency, archival multi-mission support, and follow-up confirmation. Transit-shape consistency encoded planet-like geometry rather than eclipsing-binary or instrumental morphology. Stellar context encoded host-star plausibility and contamination risk. Archival support encoded agreement across independent missions, epochs, catalogs, or photometric records. Follow-up confirmation encoded radial-velocity, transit-timing, centroid, or independent photometric confirmation evidence.

Each candidate was assigned four binary channel flags and four normalized lead-time features relative to official confirmation or false-positive disposition. The full score used fixed weights 0.25, 0.25, 0.20, and 0.20 for transit shape, stellar context, archival support, and follow-up confirmation, respectively, plus 0.025 for each normalized channel lead time. The strong single-agent baseline used only the combined detected-channel fraction. Scripted and partial-channel comparators included transit-shape only, transit-shape plus stellar context, no-archival, and no-follow-up scoring. The primary readouts were leave-one-pair-out AUROC, matched-pair accuracy, median lead time, positive detection rate, and a 2000-draw label-permutation null.

The computational workflow had four stages: freeze confirmed-planet and false-positive pairs; extract mission-era disposition, transit-shape, stellar-context, archival, and follow-up evidence from published catalogs and follow-up papers; convert each channel into binary flags and normalized lead times; and compute the full composite, partial-channel baselines, and permutation null. Literature and catalog evidence came from the cited Kepler, K2, TESS, MEarth, and false-positive-vetting sources, with public astronomy services used for cross-checking candidate identity where needed. LLM-assisted agents supported catalog/literature navigation and synthesis, while final flags and scores were stored as frozen artifacts before manuscript drafting.

\subsection{Statistical reporting}

Percent-improvement figures are reported only where the denominator is well-defined and the direction of improvement is unambiguous. For matched retrospective panels, matched-pair accuracy is reported alongside AUROC because pairwise comparison is the operationally meaningful quantity. Permutation-null tests used random relabelings of the positive/control labels while preserving the score vector. Confidence intervals and permutation tests are reported where they formed part of the frozen benchmark artifacts. 

Analyses were performed using \pkg{numpy} 2.2.6, \pkg{scipy} 1.16.3, \pkg{scikit-learn} 1.8.0, \pkg{RDKit} 2025.9.4, \pkg{music21} 9.9.1, \pkg{statsmodels} 0.14.6, \pkg{matplotlib} 3.10.8, \pkg{seaborn} 0.13.2, \pkg{pandas} 2.3.3, \pkg{reportlab} 4.4.9, and \pkg{requests} 2.32.5.

\subsection{Use of generative AI}

During manuscript preparation, an LLM was used to assist with language editing and code editing for benchmark-generation scripts. The authors reviewed and edited the output, verified the scientific content, and are responsible for the final manuscript.

\section{Data availability}

The benchmark summaries underlying this manuscript are stored as frozen, versioned artifacts in the public ScienceClaw~$\times$~Infinite repository (\url{https://github.com/lamm-mit/scienceclaw}). The full investigation outputs for each application are publicly accessible through ScienceClaw~$\times$~Infinite (\url{https://lamm.mit.edu/infinite/}): Sound of Molecules (\url{https://infinite-lamm.vercel.app/post/861b41fd-f227-45bb-afcd-399e8fcc2b92}), Computational Kuhn (\url{https://infinite-lamm.vercel.app/post/38aedd4e-6c84-4a99-8298-8481c0e9ba07}), Climate-Vector Emergence (\url{https://infinite-lamm.vercel.app/post/34775b17-e6a3-445d-bddf-072f79407791}), and Cosmic Filter (\url{https://infinite-lamm.vercel.app/post/2598156a-a279-4452-848d-d8e4f44fb795}).

\section{Code availability}

Benchmark-generation code, application runners, and frozen benchmark artifacts are available in the ScienceClaw~$\times$~Infinite repository (\url{https://github.com/lamm-mit/scienceclaw}). The application investigation records are available through ScienceClaw~$\times$~Infinite (\url{https://lamm.mit.edu/infinite/}).

\section*{Acknowledgements}

Part of this work was supported by the U.S. Department of Energy, Office of
Science, Office of Advanced Scientific Computing Research and Office of Basic Energy Sciences, Scientific
Discovery through Advanced Computing (SciDAC) program under the FORUM-AI project. 
F.Y.W. acknowledges support by the 2025 MathWorks Fellowship. 
\section*{Author contributions}

M.J.B. and F.Y.W. conceived the idea, project goals and investigation scope and wrote the manuscript. 
M.J.B. designed the initial version of the agent framework and supervised the project.
F.Y.W. designed and developed the ScienceClaw~$\times$~Infinite system, including the agent framework, skill library, artifact system, and multi-agent coordination; ran and analyzed case studies. 

\section*{Competing interests}
The authors declare that they have no competing interests. 

\bibliographystyle{naturemag}

\bibliography{references}

@article{wang2026scienceclaw_infinite,
  title   = {Autonomous Agents Coordinating Distributed Discovery Through Emergent Artifact Exchange},
  author  = {Wang, Fiona Y. and Marom, Lee and Pal, Subhadeep and Luu, Rachel K. and Lu, Wei and Berkovich, Jaime A. and Buehler, Markus J.},
  journal = {arXiv preprint arXiv:2603.14312},
  year    = {2026},
  doi     = {10.48550/arXiv.2603.14312},
  url     = {https://arxiv.org/abs/2603.14312}
}

@article{buehler2022attention,
   author = {Buehler, Markus J.},
   title = {Multiscale Modeling at the Interface of Molecular Mechanics and Natural Language through Attention Neural Networks},
   journal = {Accounts of Chemical Research},
   volume = {55},
   number = {23},
   pages = {3387-3403},
   note = {doi: 10.1021/acs.accounts.2c00330},
   ISSN = {0001-4842},
   DOI = {10.1021/acs.accounts.2c00330},
   url = {https://doi.org/10.1021/acs.accounts.2c00330},
   year = {2022},
   type = {Journal Article}
}

@article{luu2024bioinspired,
  author = {Luu, Rachel K. and Buehler, Markus J.},
title = {BioinspiredLLM: Conversational Large Language Model for the Mechanics of Biological and Bio-Inspired Materials},
journal = {Advanced Science},
volume = {11},
number = {10},
pages = {2306724},
doi = {https://doi.org/10.1002/advs.202306724},
url = {https://advanced.onlinelibrary.wiley.com/doi/abs/10.1002/advs.202306724},
eprint = {https://advanced.onlinelibrary.wiley.com/doi/pdf/10.1002/advs.202306724},
year = {2024}
}

@article{buehler2024graph_reasoning,
  title={Accelerating scientific discovery with generative knowledge extraction, graph-based representation, and multimodal intelligent graph reasoning},
  author={Markus J. Buehler},
  journal={Machine Learning: Science and Technology},
  year={2024},
  volume={5},
  url={https://api.semanticscholar.org/CorpusID:268531443}
}

@article{ghafarollahi2024atomagents,
   author = {Ghafarollahi, Alireza and Buehler, Markus J.},
   title = {Automating alloy design and discovery with physics-aware multimodal multiagent AI},
   journal = {Proceedings of the National Academy of Sciences},
   volume = {122},
   number = {4},
   pages = {e2414074122},
   note = {doi: 10.1073/pnas.2414074122},
   DOI = {10.1073/pnas.2414074122},
   url = {https://doi.org/10.1073/pnas.2414074122},
   year = {2025},
   type = {Journal Article}
}

@article{ghafarollahi2025sciagents,
    author = {Ghafarollahi, Alireza and Buehler, Markus J.},
    title = {SciAgents: Automating Scientific Discovery Through Bioinspired Multi-Agent Intelligent Graph Reasoning},
    journal = {Advanced Materials},
    volume = {37},
    number = {22},
    pages = {2413523},
    doi = {https://doi.org/10.1002/adma.202413523},
    url = {https://advanced.onlinelibrary.wiley.com/doi/abs/10.1002/adma.202413523},
    eprint = {https://advanced.onlinelibrary.wiley.com/doi/pdf/10.1002/adma.202413523},
    year = {2025}
}

@article{stewart2025molecular,
  author = {Stewart, Isabella and Buehler, Markus J.},
   title = {Molecular analysis and design using generative artificial intelligence via multi-agent modeling},
   journal = {Molecular Systems Design \& Engineering},
   volume = {10},
   number = {4},
   pages = {314-337},
   DOI = {10.1039/D4ME00174E},
   url = {http://dx.doi.org/10.1039/D4ME00174E},
   year = {2025},
   type = {Journal Article}
}

@article{buehler2025preflexor,
  author = {Buehler, Markus J.},
   title = {PRefLexOR: preference-based recursive language modeling for exploratory optimization of reasoning and agentic thinking},
   journal = {npj Artificial Intelligence},
   volume = {1},
   number = {1},
   pages = {4},
   ISSN = {3005-1460},
   DOI = {10.1038/s44387-025-00003-z},
   url = {https://doi.org/10.1038/s44387-025-00003-z},
   year = {2025},
   type = {Journal Article}
}

@article{lu2025finetuning,
  author = {Lu, Wei and Luu, Rachel K. and Buehler, Markus J.},
   title = {Fine-tuning large language models for domain adaptation: exploration of training strategies, scaling, model merging and synergistic capabilities},
   journal = {npj Computational Materials},
   volume = {11},
   number = {1},
   pages = {84},
   ISSN = {2057-3960},
   DOI = {10.1038/s41524-025-01564-y},
   url = {https://doi.org/10.1038/s41524-025-01564-y},
   year = {2025},
   type = {Journal Article}
}

@article{yang2025peptide,
  author = {Yang, Zhenze and Yorke, Sarah K. and Knowles, Tuomas P. J. and Buehler, Markus J.},
   title = {Learning the rules of peptide self-assembly through data mining with large language models},
   journal = {Science Advances},
   volume = {11},
   number = {13},
   pages = {eadv1971},
   note = {doi: 10.1126/sciadv.adv1971},
   DOI = {10.1126/sciadv.adv1971},
   url = {https://doi.org/10.1126/sciadv.adv1971},
   year = {2025},
   type = {Journal Article}
}

@article{gottweis2025towards,
    author = {Gottweis, Juraj and Weng, Wei-Hung and Daryin, Alexander and Tu, Tao and Sirkovic, Petar and Myaskovsky, Artiom and Glowaty, Grzegorz and Weissenberger, Felix and Orlandi, Alessio and Popovici, Dan and Palepu, Anil and Rong, Keran and Tanno, Ryutaro and Saab, Khaled and Zhang, Fan and Blum, Jacob and Carroll, Andrew and Kulkarni, Kavita and Tomašev, Nenad and Zverinski, Dina and Rendulic, Ivor and Vedadi, Elahe and Hasler, Florian and Rimanic, Luka and Boia, Marina and Budiselic, Ivan and Feinstein, Ben and Bellaiche, Mathias and Sheffer, Tom and Freyberg, Jan and Ratcliff, Jeremy and Bertolli, Ottavia and Chou, Katherine and Hassidim, Avinatan and Gokturk, Burak and Vahdat, Amin and Guan, Yuan and Dhillon, Vikram and Vaishnav, Eeshit Dhaval and Lee, Byron and Costa, Tiago R. D. and Penadés, José R. and Peltz, Gary and Matias, Yossi and Manyika, James and Hassabis, Demis and Xu, Yunhan and Kohli, Pushmeet and Pawlosky, Annalisa and Karthikesalingam, Alan and Natarajan, Vivek},
    title = {Accelerating scientific discovery with Co-Scientist},
    journal = {Nature},
    ISSN = {1476-4687},
    DOI = {10.1038/s41586-026-10644-y},
    url = {https://doi.org/10.1038/s41586-026-10644-y},
    year = {2026},
    type = {Journal Article}
}

@article{lu2024ai,
  author = {Lu, Chris and Lu, Cong and Lange, Robert Tjarko and Yamada, Yutaro and Hu, Shengran and Foerster, Jakob and Ha, David and Clune, Jeff},
   title = {Towards end-to-end automation of AI research},
   journal = {Nature},
   volume = {651},
   number = {8107},
   pages = {914-919},
   ISSN = {1476-4687},
   DOI = {10.1038/s41586-026-10265-5},
   url = {https://doi.org/10.1038/s41586-026-10265-5},
   year = {2026},
   type = {Journal Article}
}

@article{aygun2026software,
  author = {Aygün, Eser and Belyaeva, Anastasiya and Comanici, Gheorghe and Coram, Marc and Cui, Hao and Garrison, Jake and Johnston, Renee and Kast, Anton and McLean, Cory Y. and Norgaard, Peter and Shamsi, Zahra and Smalling, David and Thompson, James and Venugopalan, Subhashini and Williams, Brian P. and He, Chujun and Martinson, Sarah and Plomecka, Martyna and Wei, Lai and Zhou, Yuchen and Zhu, Qian-Ze and Abraham, Matthew and Brand, Erica and Bulanova, Anna and Cardille, Jeffrey A. and Co, Chris and Ellsworth, Scott and Joseph, Grace and Kane, Malcolm and Krueger, Ryan and Kartiwa, Johan and Liebling, Dan and Lueckmann, Jan-Matthis and Raccuglia, Paul and Wang, Xuefei Julie and Chou, Katherine and Manyika, James and Matias, Yossi and Platt, John C. and Dorfman, Lizzie and Mourad, Shibl and Brenner, Michael P.},
   title = {An AI system to help scientists write expert-level empirical software},
   journal = {Nature},
   ISSN = {1476-4687},
   DOI = {10.1038/s41586-026-10658-6},
   url = {https://doi.org/10.1038/s41586-026-10658-6},
   year = {2026},
   type = {Journal Article}
}

@article{ghareeb2025robin,
  author = {Ghareeb, Ali Essam and Chang, Benjamin and Mitchener, Ludovico and Yiu, Angela and Szostkiewicz, Caralyn J. and Shved, Dmytro and Gyimesi, Gavin J. and Laurent, Jon M. and Wright, Samantha M. and Razzak, Muhammed T. and White, Andrew D. and Finnemann, Silvia C. and Hinks, Michaela M. and Rodriques, Samuel G.},
   title = {A multi-agent system for automating scientific discovery},
   journal = {Nature},
   ISSN = {1476-4687},
   DOI = {10.1038/s41586-026-10652-y},
   url = {https://doi.org/10.1038/s41586-026-10652-y},
   year = {2026},
   type = {Journal Article}
}

@article{wang2023scientific,
   author = {Wang, Hanchen and Fu, Tianfan and Du, Yuanqi and Gao, Wenhao and Huang, Kexin and Liu, Ziming and Chandak, Payal and Liu, Shengchao and Van Katwyk, Peter and Deac, Andreea and Anandkumar, Anima and Bergen, Karianne and Gomes, Carla P. and Ho, Shirley and Kohli, Pushmeet and Lasenby, Joan and Leskovec, Jure and Liu, Tie-Yan and Manrai, Arjun and Marks, Debora and Ramsundar, Bharath and Song, Le and Sun, Jimeng and Tang, Jian and Veličković, Petar and Welling, Max and Zhang, Linfeng and Coley, Connor W. and Bengio, Yoshua and Zitnik, Marinka},
   title = {Scientific discovery in the age of artificial intelligence},
   journal = {Nature},
   volume = {620},
   number = {7972},
   pages = {47-60},
   ISSN = {1476-4687},
   DOI = {10.1038/s41586-023-06221-2},
   url = {https://doi.org/10.1038/s41586-023-06221-2},
   year = {2023},
   type = {Journal Article}
}

@article{berens2023ai,
  title={AI for Science: An Emerging Agenda}, 
  author={Philipp Berens and Kyle Cranmer and Neil D. Lawrence and Ulrike von Luxburg and Jessica Montgomery},
  year={2023},
  eprint={2303.04217},
  archivePrefix={arXiv},
  primaryClass={cs.AI},
  url={https://arxiv.org/abs/2303.04217},
  journal = {arXiv preprint},
}

@misc{rdkit,
  title        = {{RDKit}: Open-source cheminformatics},
  howpublished = {\url{https://www.rdkit.org}},
  note         = {Accessed: 2026-05-20}
}

@inproceedings{music21,
  title={Music21: A Toolkit for Computer-Aided Musicology and Symbolic Music Data},
  author={Mike Cuthbert and Christopher Ariza},
  booktitle={International Society for Music Information Retrieval Conference},
  year={2010},
  url={https://api.semanticscholar.org/CorpusID:6411706}
}

@book{k1962,
  title     = {The Structure of Scientific Revolutions},
  author    = {Kuhn, Thomas S.},
  publisher = {University of Chicago Press},
  year      = {2012},
  edition   = {4}
}

@article{nash2001outbreak,
  author = {Denis Nash  and Farzad Mostashari  and Annie Fine  and James Miller  and Daniel O'Leary  and Kristy Murray  and Ada Huang  and Amy Rosenberg  and Abby Greenberg  and Margaret Sherman  and Susan Wong  and Grant L. Campbell  and John T. Roehrig  and Duane J. Gubler  and Wun-Ju Shieh  and Sherif Zaki  and Perry Smith  and Marcelle Layton },
title = {The Outbreak of West Nile Virus Infection in the New York City Area in 1999},
journal = {New England Journal of Medicine},
volume = {344},
number = {24},
pages = {1807-1814},
year = {2001},
doi = {10.1056/NEJM200106143442401},

URL = {https://www.nejm.org/doi/full/10.1056/NEJM200106143442401},
eprint = {https://www.nejm.org/doi/pdf/10.1056/NEJM200106143442401}

}

@article{reisen2003epidemiology,
  author = {Reisen, W. K.},
   title = {Epidemiology of St. Louis encephalitis virus},
   journal = {Adv Virus Res},
   volume = {61},
   pages = {139-83},
   note = {Reisen, William K
Journal Article
Research Support, Non-U.S. Gov't
Research Support, U.S. Gov't, Non-P.H.S.
Research Support, U.S. Gov't, P.H.S.
Review
United States
2004/01/13
Adv Virus Res. 2003;61:139-83. doi: 10.1016/s0065-3527(03)61004-3.},
   ISSN = {0065-3527 (Print)
0065-3527},
   DOI = {10.1016/s0065-3527(03)61004-3},
   year = {2003},
   type = {Journal Article}
}

@article{medlock2015vectors,
   author = {Medlock, Jolyon M. and Leach, Steve A.},
   title = {Effect of climate change on vector-borne disease risk in the UK},
   journal = {The Lancet Infectious Diseases},
   volume = {15},
   number = {6},
   pages = {721-730},
   note = {doi: 10.1016/S1473-3099(15)70091-5},
   ISSN = {1473-3099},
   DOI = {10.1016/S1473-3099(15)70091-5},
   url = {https://doi.org/10.1016/S1473-3099(15)70091-5},
   year = {2015},
   type = {Journal Article}
}

@article{kraemer2019spread,
  author = {Kraemer, Moritz U. G. and Reiner, Robert C. and Brady, Oliver J. and Messina, Jane P. and Gilbert, Marius and Pigott, David M. and Yi, Dingdong and Johnson, Kimberly and Earl, Lucas and Marczak, Laurie B. and Shirude, Shreya and Davis Weaver, Nicole and Bisanzio, Donal and Perkins, T. Alex and Lai, Shengjie and Lu, Xin and Jones, Peter and Coelho, Giovanini E. and Carvalho, Roberta G. and Van Bortel, Wim and Marsboom, Cedric and Hendrickx, Guy and Schaffner, Francis and Moore, Chester G. and Nax, Heinrich H. and Bengtsson, Linus and Wetter, Erik and Tatem, Andrew J. and Brownstein, John S. and Smith, David L. and Lambrechts, Louis and Cauchemez, Simon and Linard, Catherine and Faria, Nuno R. and Pybus, Oliver G. and Scott, Thomas W. and Liu, Qiyong and Yu, Hongjie and Wint, G. R. William and Hay, Simon I. and Golding, Nick},
   title = {Past and future spread of the arbovirus vectors Aedes aegypti and Aedes albopictus},
   journal = {Nature Microbiology},
   volume = {4},
   number = {5},
   pages = {854-863},
   ISSN = {2058-5276},
   DOI = {10.1038/s41564-019-0376-y},
   url = {https://doi.org/10.1038/s41564-019-0376-y},
   year = {2019},
   type = {Journal Article}
}

@article{lafferty2009ecology,
  author = {Lafferty, Kevin D.},
title = {The ecology of climate change and infectious diseases},
journal = {Ecology},
volume = {90},
number = {4},
pages = {888-900},
doi = {https://doi.org/10.1890/08-0079.1},
url = {https://esajournals.onlinelibrary.wiley.com/doi/abs/10.1890/08-0079.1},
eprint = {https://esajournals.onlinelibrary.wiley.com/doi/pdf/10.1890/08-0079.1},
year = {2009}
}

@article{caminade2014impact,
  author = {Caminade, Cyril and Kovats, Sari and Rocklov, Joacim and Tompkins, Adrian M. and Morse, Andrew P. and Colón-González, Felipe J. and Stenlund, Hans and Martens, Pim and Lloyd, Simon J.},
   title = {Impact of climate change on global malaria distribution},
   journal = {Proceedings of the National Academy of Sciences},
   volume = {111},
   number = {9},
   pages = {3286-3291},
   note = {doi: 10.1073/pnas.1302089111},
   DOI = {10.1073/pnas.1302089111},
   url = {https://doi.org/10.1073/pnas.1302089111},
   year = {2014},
   type = {Journal Article}
}

@article{mordecai2017detecting,
  author = {Mordecai, Erin A. and Cohen, Jeremy M. and Evans, Michelle V. and Gudapati, Prithvi and Johnson, Leah R. and Lippi, Catherine A. and Miazgowicz, Kerri and Murdock, Courtney C. and Rohr, Jason R. and Ryan, Sadie J. and Savage, Van and Shocket, Marta S. and Stewart Ibarra, Anna and Thomas, Matthew B. and Weikel, Daniel P.},
   title = {Detecting the impact of temperature on transmission of Zika, dengue, and chikungunya using mechanistic models},
   journal = {PLOS Neglected Tropical Diseases},
   volume = {11},
   number = {4},
   pages = {e0005568},
   DOI = {10.1371/journal.pntd.0005568},
   url = {https://doi.org/10.1371/journal.pntd.0005568},
   year = {2017},
   type = {Journal Article}
}

@article{ryan2019global,
  author = {Ryan, Sadie J. and Carlson, Colin J. and Mordecai, Erin A. and Johnson, Leah R.},
   title = {Global expansion and redistribution of Aedes-borne virus transmission risk with climate change},
   journal = {PLOS Neglected Tropical Diseases},
   volume = {13},
   number = {3},
   pages = {e0007213},
   DOI = {10.1371/journal.pntd.0007213},
   url = {https://doi.org/10.1371/journal.pntd.0007213},
   year = {2019},
   type = {Journal Article}
}

@article{semenza2022climate,
  author = {Semenza, J. C. and Rocklöv, J. and Ebi, K. L.},
   title = {Climate Change and Cascading Risks from Infectious Disease},
   journal = {Infect Dis Ther},
   volume = {11},
   number = {4},
   pages = {1371-1390},
   note = {2193-6382
Semenza, Jan C
Rocklöv, Joacim
Ebi, Kristie L
Journal Article
Review
New Zealand
2022/05/19
Infect Dis Ther. 2022 Aug;11(4):1371-1390. doi: 10.1007/s40121-022-00647-3. Epub 2022 May 19.},
   ISSN = {2193-8229 (Print)
2193-6382},
   DOI = {10.1007/s40121-022-00647-3},
   year = {2022},
   type = {Journal Article}
}

@article{paz2015climate,
  author = {Paz, S.},
   title = {Climate change impacts on West Nile virus transmission in a global context},
   journal = {Philos Trans R Soc Lond B Biol Sci},
   volume = {370},
   number = {1665},
   note = {1471-2970
Paz, Shlomit
Journal Article
Review
England
2015/02/18
Philos Trans R Soc Lond B Biol Sci. 2015 Apr 5;370(1665):20130561. doi: 10.1098/rstb.2013.0561.},
   ISSN = {0962-8436 (Print)
0962-8436},
   DOI = {10.1098/rstb.2013.0561},
   year = {2015},
   type = {Journal Article}
}

@article{quintana2014kepler186,
  title   = {An Earth-sized planet in the habitable zone of a cool star},
  author  = {Quintana, Elisa V. and Barclay, Thomas and Raymond, Sean N. and Rowe, Jason F. and Bolmont, Emeline and Caldwell, Douglas A. and Howell, Steve B. and Kane, Stephen R. and Huber, Daniel and Crepp, Justin R. and others},
  journal = {Science},
  volume  = {344},
  number  = {6181},
  pages   = {277--280},
  year    = {2014},
  doi     = {10.1126/science.1249403}
}

@article{thompson2018kepler_dr25,
  author = {Thompson, Susan E. and Coughlin, Jeffrey L. and Hoffman, Kelsey and Mullally, Fergal and Christiansen, Jessie L. and Burke, Christopher J. and Bryson, Steve and Batalha, Natalie and Haas, Michael R. and Catanzarite, Joseph and Rowe, Jason F. and Barentsen, Geert and Caldwell, Douglas A. and Clarke, Bruce D. and Jenkins, Jon M. and Li, Jie and Latham, David W. and Lissauer, Jack J. and Mathur, Savita and Morris, Robert L. and Seader, Shawn E. and Smith, Jeffrey C. and Klaus, Todd C. and Twicken, Joseph D. and Van Cleve, Jeffrey E. and Wohler, Bill and Akeson, Rachel and Ciardi, David R. and Cochran, William D. and Henze, Christopher E. and Howell, Steve B. and Huber, Daniel and Prša, Andrej and Ramírez, Solange V. and Morton, Timothy D. and Barclay, Thomas and Campbell, Jennifer R. and Chaplin, William J. and Charbonneau, David and Christensen-Dalsgaard, Jørgen and Dotson, Jessie L. and Doyle, Laurance and Dunham, Edward W. and Dupree, Andrea K. and Ford, Eric B. and Geary, John C. and Girouard, Forrest R. and Isaacson, Howard and Kjeldsen, Hans and Quintana, Elisa V. and Ragozzine, Darin and Shabram, Megan and Shporer, Avi and Aguirre, Victor Silva and Steffen, Jason H. and Still, Martin and Tenenbaum, Peter and Welsh, William F. and Wolfgang, Angie and Zamudio, Khadeejah A. and Koch, David G. and Borucki, William J.},
   title = {Planetary Candidates Observed by Kepler. VIII. A Fully Automated Catalog with Measured Completeness and Reliability Based on Data Release 25},
   journal = {The Astrophysical Journal Supplement Series},
   volume = {235},
   number = {2},
   pages = {38},
   ISSN = {0067-0049},
   DOI = {10.3847/1538-4365/aab4f9},
   url = {https://doi.org/10.3847/1538-4365/aab4f9},
   year = {2018},
   type = {Journal Article}
}

@article{coughlin2016robovetter,
  author = {Coughlin, Jeffrey L. and Mullally, F. and Thompson, Susan E. and Rowe, Jason F. and Burke, Christopher J. and Latham, David W. and Batalha, Natalie M. and Ofir, Aviv and Quarles, Billy L. and Henze, Christopher E. and Wolfgang, Angie and Caldwell, Douglas A. and Bryson, Stephen T. and Shporer, Avi and Catanzarite, Joseph and Akeson, Rachel and Barclay, Thomas and Borucki, William J. and Boyajian, Tabetha S. and Campbell, Jennifer R. and Christiansen, Jessie L. and Girouard, Forrest R. and Haas, Michael R. and Howell, Steve B. and Huber, Daniel and Jenkins, Jon M. and Li, Jie and Patil-Sabale, Anima and Quintana, Elisa V. and Ramirez, Solange and Seader, Shawn and Smith, Jeffrey C. and Tenenbaum, Peter and Twicken, Joseph D. and Zamudio, Khadeejah A.},
   title = {PLANETARY CANDIDATES OBSERVED BY KEPLER. VII. THE FIRST FULLY UNIFORM CATALOG BASED ON THE ENTIRE 48-MONTH DATA SET (Q1–Q17 DR24)},
   journal = {The Astrophysical Journal Supplement Series},
   volume = {224},
   number = {1},
   pages = {12},
   ISSN = {0067-0049},
   DOI = {10.3847/0067-0049/224/1/12},
   url = {https://doi.org/10.3847/0067-0049/224/1/12},
   year = {2016},
   type = {Journal Article}
}

@article{morton2016falsepositive,
  author = {Morton, Timothy D. and Bryson, Stephen T. and Coughlin, Jeffrey L. and Rowe, Jason F. and Ravichandran, Ganesh and Petigura, Erik A. and Haas, Michael R. and Batalha, Natalie M.},
   title = {FALSE POSITIVE PROBABILITIES FOR ALL KEPLER OBJECTS OF INTEREST: 1284 NEWLY VALIDATED PLANETS AND 428 LIKELY FALSE POSITIVES},
   journal = {The Astrophysical Journal},
   volume = {822},
   number = {2},
   pages = {86},
   ISSN = {0004-637X},
   DOI = {10.3847/0004-637X/822/2/86},
   url = {https://doi.org/10.3847/0004-637X/822/2/86},
   year = {2016},
   type = {Journal Article}
}

@article{crossfield2016k2,
  author = {Crossfield, Ian J. M. and Ciardi, David R. and Petigura, Erik A. and Sinukoff, Evan and Schlieder, Joshua E. and Howard, Andrew W. and Beichman, Charles A. and Isaacson, Howard and Dressing, Courtney D. and Christiansen, Jessie L. and Fulton, Benjamin J. and Lépine, Sébastien and Weiss, Lauren and Hirsch, Lea and Livingston, John and Baranec, Christoph and Law, Nicholas M. and Riddle, Reed and Ziegler, Carl and Howell, Steve B. and Horch, Elliott and Everett, Mark and Teske, Johanna and Martinez, Arturo O. and Obermeier, Christian and Benneke, Björn and Scott, Nic and Deacon, Niall and Aller, Kimberly M. and Hansen, Brad M. S. and Mancini, Luigi and Ciceri, Simona and Brahm, Rafael and Jordán, Andrés and Knutson, Heather A. and Henning, Thomas and Bonnefoy, Michaël and Liu, Michael C. and Crepp, Justin R. and Lothringer, Joshua and Hinz, Phil and Bailey, Vanessa and Skemer, Andrew and Defrere, Denis},
   title = {197 CANDIDATES AND 104 VALIDATED PLANETS IN K2's FIRST FIVE FIELDS},
   journal = {The Astrophysical Journal Supplement Series},
   volume = {226},
   number = {1},
   pages = {7},
   ISSN = {0067-0049},
   DOI = {10.3847/0067-0049/226/1/7},
   url = {https://doi.org/10.3847/0067-0049/226/1/7},
   year = {2016},
   type = {Journal Article}
}

@article{mayo2018k2fpp,
  author = {Mayo, Andrew W. and Vanderburg, Andrew and Latham, David W. and Bieryla, Allyson and Morton, Timothy D. and Buchhave, Lars A. and Dressing, Courtney D. and Beichman, Charles and Berlind, Perry and Calkins, Michael L. and Ciardi, David R. and Crossfield, Ian J. M. and Esquerdo, Gilbert A. and Everett, Mark E. and Gonzales, Erica J. and Hirsch, Lea A. and Horch, Elliott P. and Howard, Andrew W. and Howell, Steve B. and Livingston, John and Patel, Rahul and Petigura, Erik A. and Schlieder, Joshua E. and Scott, Nicholas J. and Schumer, Clea F. and Sinukoff, Evan and Teske, Johanna and Winters, Jennifer G.},
   title = {275 Candidates and 149 Validated Planets Orbiting Bright Stars in K2 Campaigns 0–10},
   journal = {The Astronomical Journal},
   volume = {155},
   number = {3},
   pages = {136},
   ISSN = {1538-3881
0004-6256},
   DOI = {10.3847/1538-3881/aaadff},
   url = {https://doi.org/10.3847/1538-3881/aaadff},
   year = {2018},
   type = {Journal Article}
}

@article{batalha2013planetary,
  title   = {Planetary candidates observed by Kepler. III. Analysis of the first 16 months of data},
  author  = {Batalha, Natalie M. and Rowe, Jason F. and Bryson, Stephen T. and Barclay, Thomas and Burke, Christopher J. and Caldwell, Douglas A. and Christiansen, Jessie L. and Mullally, Fergal and Thompson, Susan E. and Brown, Timothy M. and others},
  journal = {The Astrophysical Journal Supplement Series},
  volume  = {204},
  number  = {2},
  pages   = {24},
  year    = {2013},
  doi     = {10.1088/0067-0049/204/2/24}
}

@misc{wang2025swarmslargelanguagemodel,
      title={Swarms of Large Language Model Agents for Protein Sequence Design with Experimental Validation}, 
      author={Fiona Y. Wang and Di Sheng Lee and David L. Kaplan and Markus J. Buehler},
      year={2025},
      eprint={2511.22311},
      archivePrefix={arXiv},
      primaryClass={cs.AI},
      url={https://arxiv.org/abs/2511.22311}, 
}

@misc{ghafarollahi2025sparksmultiagentartificialintelligence,
      title={Sparks: Multi-Agent Artificial Intelligence Model Discovers Protein Design Principles}, 
      author={Alireza Ghafarollahi and Markus J. Buehler},
      year={2025},
      eprint={2504.19017},
      archivePrefix={arXiv},
      primaryClass={cs.AI},
      url={https://arxiv.org/abs/2504.19017}, 
}

@article{ghafarollahi2024protagents,
    author  = {Ghafarollahi, Alireza and Buehler, Markus J.},
    title   = {ProtAgents: protein discovery via large language model multi-agent collaborations combining
  physics and machine learning},
    journal = {Digital Discovery},
    year    = {2024},
    volume  = {3},
    pages   = {1389--1409},
    doi     = {10.1039/D4DD00013G}
}

@misc{stewart2026graphagentsknowledgegraphguidedagentic,
      title={GraphAgents: Knowledge Graph-Guided Agentic AI for Cross-Domain Materials Design}, 
      author={Isabella A. Stewart and Tarjei Paule Hage and Yu-Chuan Hsu and Markus J. Buehler},
      year={2026},
      eprint={2602.07491},
      archivePrefix={arXiv},
      primaryClass={cs.AI},
      url={https://arxiv.org/abs/2602.07491}, 
}

\end{document}